\documentclass[runningheads]{llncs}

\usepackage{eccv}

\usepackage{eccvabbrv}

\usepackage{supertabular}
\usepackage{graphicx}
\usepackage{booktabs}
\usepackage{amsmath}

\usepackage[accsupp]{axessibility}  

\usepackage[pagebackref,breaklinks,colorlinks,citecolor=eccvblue]{hyperref}

\usepackage[utf8]{inputenc} 
\usepackage[T1]{fontenc}    
\usepackage[pagebackref,breaklinks,colorlinks]{hyperref}
\usepackage{url}      

\usepackage{supertabular}
\usepackage{amsfonts}       
\usepackage{nicefrac}       
\usepackage{microtype}      

\usepackage{tcolorbox}
\usepackage{times}
\usepackage{epsfig}
\usepackage{graphicx}
\usepackage{amsmath}
\usepackage{amssymb}
\usepackage{booktabs}
\usepackage{xcolor}

\usepackage{CJKutf8}

\usepackage{stfloats}

\newcommand{\xmark}{\textcolor{red}{\text{\ding{55}}}}
\newcommand{\cmark}{\textcolor{green}{\text{\ding{51}}}}

\usepackage{mdframed}
\usepackage{xr-hyper}
\usepackage{lscape}
\usepackage{caption}
\usepackage{subcaption}
\usepackage{multirow}
\usepackage{algorithmicx}
\usepackage{algpseudocode}
\usepackage{siunitx}
\usepackage{lipsum} 
\usepackage{caption} 

\newmdenv[
  innerleftmargin=10pt,
  innerrightmargin=10pt,
  innertopmargin=10pt,
  innerbottommargin=10pt,
  linecolor=black,
  linewidth=1pt,
  topline=true,
  bottomline=true,
  rightline=true,
  leftline=true,
  backgroundcolor=gray!20,
  nobreak=true,
]{promptbox}

\usepackage{amsmath}
\usepackage{amsfonts}
\usepackage{amssymb}
\usepackage{wrapfig}
\usepackage{subcaption}
\usepackage{multirow}
 \usepackage{mathtools} 

\usepackage{verbatim}

\usepackage{anyfontsize}

\usepackage{microtype}
\usepackage{graphicx}
\usepackage{booktabs} 
\usepackage{multirow}
\usepackage{amsmath,amssymb}
\usepackage{booktabs}
\usepackage{caption,subcaption}

\usepackage{xcolor}
\definecolor{mygreen}{HTML}{3cb44b}
\definecolor{skyblue}{HTML}{beffff}
\definecolor{lightgreen}{HTML}{90ee90}

\usepackage{color, colortbl}

\definecolor{emerald}{rgb}{0.31, 0.78, 0.37}

\usepackage{tcolorbox}
\usepackage{enumitem}
\setitemize{itemsep=10pt,topsep=0pt,parsep=0pt,partopsep=0pt}
\usepackage{colortbl}

\usepackage{xcolor}
\definecolor{mygreen}{HTML}{3cb44b}
\colorlet{myyellow}{green!10!orange!90!}
\makeatletter

\usepackage{tikz}
\usetikzlibrary{arrows,shapes,snakes,automata,backgrounds,fit,petri}
\usepackage{adjustbox}

\newcommand{\RN}[1]{%
	\textup{\lowercase\expandafter{\it \romannumeral#1}}%
}
\usepackage{tabu}

\newcommand{\beq}{\vspace{0mm}\begin{equation}}
\newcommand{\eeq}{\vspace{0mm}\end{equation}}
\newcommand{\beqs}{\vspace{0mm}\begin{eqnarray}}
\newcommand{\eeqs}{\vspace{0mm}\end{eqnarray}}
\newcommand{\barr}{\begin{array}}
\newcommand{\earr}{\end{array}}

\usepackage{color, colortbl}
\definecolor{Gray}{gray}{0.93}

\usepackage{lipsum}

\usepackage{pifont}

\usepackage{makecell}

\usepackage{xcolor,amsmath}
\usepackage[linesnumbered,ruled,vlined]{algorithm2e}
\DontPrintSemicolon

\usepackage{xcolor}
\definecolor{mygreen}{HTML}{3cb44b}

\SetKwComment{Comment}{\color{green!50!black}\# }{}

\SetKwProg{Function}{def}{:}{}

\SetKwProg{For}{for}{:}{}
\SetKwProg{If}{if}{:}{}

\usepackage[pagebackref,breaklinks,colorlinks]{hyperref}

\usepackage[capitalize]{cleveref}
\crefname{section}{Sec.}{Secs.}
\Crefname{section}{Section}{Sections}
\Crefname{table}{Table}{Tables}
\crefname{table}{Tab.}{Tabs.}

\usepackage{orcidlink}
\usepackage{arydshln} 

\begin{document}

\title{Fully Authentic Visual Question Answering Dataset \\ from Online Communities}


\author{Chongyan Chen\inst{1}\orcidlink{0000-0003-3188-6652} \and
Mengchen Liu\inst{2}\and
Noel Codella\inst{2}\and
Yunsheng Li\inst{2}\and
Lu Yuan\inst{2}\and
Danna Gurari\inst{1,3}}

\authorrunning{Chen, C., Liu, M., et al.}

\institute{$~^1$ University of Texas at Austin, $~^2$ Microsoft, $~^3$ University of Colorado Boulder}

\maketitle
\begin{abstract}
Visual Question Answering (VQA) entails answering questions about images.  We introduce the first VQA dataset in which all contents originate from an authentic use case.  Sourced from online question answering community forums, we call it VQAonline.  We characterize this dataset and how it relates to eight mainstream VQA datasets.  Observing that answers in our dataset tend to be much longer (i.e., a mean of 173 words) and so incompatible with standard VQA evaluation metrics, we instead utilize popular metrics for longer text evaluation for evaluating six state-of-the-art VQA models on VQAonline and report where they struggle most.  Finally, we analyze which evaluation metrics align best with human judgments. We publicly-share the dataset at: https://vqaonline.github.io/.
\end{abstract}

\section{Introduction} 
We are witnessing a renewed excitement in visual question answering (VQA), the task of answering questions about images, with recent successes of large foundation models in zero-shot and few-shot settings.  Moreover, many papers are reporting that such models can achieve or even exceed human performance~\cite{yang2023dawn,lu2023mathvista, OpenAI2023GPT4TR}.

While modern VQA datasets enabled this exciting progress in VQA model development, they have two key limitations.  First, most publicly-available VQA datasets~\cite{balanced_vqa_v2,singh2019towards,saikh2022scienceqa} are artificially-constructed to probe model performance rather stemming from authentic use cases.  Yet, repeatedly, prior work~\cite{gurari2018vizwiz,zeng2020vision,gurari2020captioning} has revealed that models developed for artificially-constructed settings generalize poorly to authentic use cases due to significant domain shifts.  The second key limitation of existing datasets lies in a limited diversity of content representing authentic use cases.  The only VQA dataset reflecting an authentic use case is VizWiz-VQA~\cite{gurari2017crowdverge}.  However, VizWiz-VQA is constrained to one population, specifically blind individuals seeking assistance in learning about their visual surroundings.  A further limitation of the dataset is that the answers were collected years later to support model benchmarking, rather than having been validated as acceptable by those who asked the questions. 

\begin{figure}[t]
     \centering
     \includegraphics[width=1\textwidth]{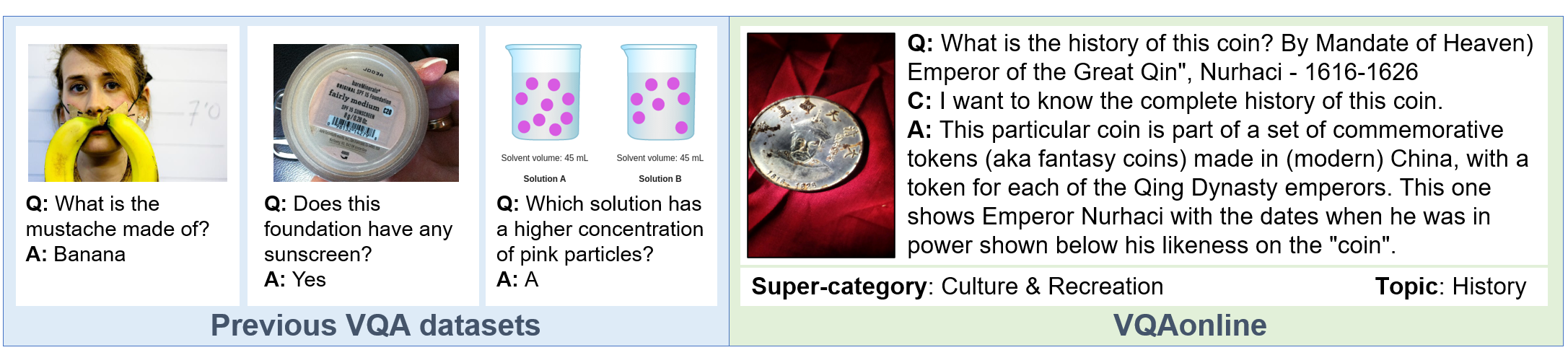}
        \caption{VQA examples from our VQAonline dataset and three mainstream VQA datasets~\cite{balanced_vqa_v2,singh2019towards,saikh2022scienceqa}.  VQAonline is the first VQA dataset to originate from an \emph{authentic} use case end to end, including with authentic context, answers, and topics/categories labels. It also is the first VQA dataset from an online question answering platform.  A critical distinction of VQAonline, which necessitates a new evaluation methodology, is its lengthy answers. (Q=question, C=context, A=answer).  }
    \label{fig:mainFigure}
\end{figure}

We address these limitations by introducing a new dataset called \emph{VQAonline.} Sourced from online question answering communities, it is the the first VQA dataset with all contents originating from an authentic use case.  We conduct analysis to highlight how this dataset differs from prior datasets; examples include that it contains: (1) authentic context that clarifies the question, (2) an answer the individual asking the question validated as acceptable, (3) answers that are considerably longer (\ie, a mean of 173 words versus typically 11 words or fewer in prior work), and (4) user-chosen topics for each visual question from 105 diverse topics.  Such differences are exemplified in Figure~\ref{fig:mainFigure}.  Next, observing that the mainstream VQA evaluation metric is ill-suited our dataset's lengthy textual answers, we benchmarked the performance of six modern foundation models with respect to popular evaluation metrics for long-form text. Our findings reveal that modern models perform poorly overall, underscoring of this dataset to facilitate the research community's continued improvement of VQA models. We conclude with assessing which related evaluation metrics for long-form text best align with human judgement for our dataset to also facilitate future work in improving evaluation metrics for lengthy textual answers. 

We expect improved models designed for VQAonline can also benefit many other real-world applications including for other community question answering platforms (\eg, Yahoo! Answers, Wiki Answers, Quora), learning assistance VQA tools for children \cite{he2017educational}, virtual assistants, visual dialog and query, as well as visual search and browsing. To encourage community-wide progress, we publicly-share the dataset at https://vqaonline.github.io/.
\section{Related Work}

\paragraph{\bf{Visual Question Answering Datasets.}}
A commonality of all VQA datasets (\eg, ~\cite{antol2015vqa,balanced_vqa_v2,gurari2018vizwiz,singh2019towards,johnson2017clevr,hudson2019gqa,chen2023can})  is the inclusion of artificially-generated content.  The closest dataset to reflect an authentic use case is VizWiz-VQA~\cite{gurari2018vizwiz}, as the visual questions were all submitted by individuals seeking assistance to learn about their visual surroundings.  However, the answers were crowdsourced and never shared with or validated by those asking the questions.  To our knowledge, our work is the first to provide an authentic dataset end-to-end, with each example including a natural language question, image, context, and answer validated by the individual asking the question.  We conduct extensive analysis to reveal unique aspects of our dataset (Section~\ref{sec:dataset_analysis}).

Our work also aligns with VQA datasets providing context that supplements each visual question: Context-VQA~\cite{naik2023context} and ScienceQA~\cite{saikh2022scienceqa}. However, \emph{context} in Context-VQA \cite{naik2023context} is defined as the kind of website the image is sourced from rather than our definition of supplementary data that supports answering the visual question.  More similar to our work is ScienceQA~\cite{saikh2022scienceqa}, however the context is both artificially-generated and has distinct compositions and content, as discussed in Section~\ref{sec:dataset_analysis}.

\paragraph{\bf{Community Question Answering (CQA) Datasets.}}
Numerous datasets have been developed around CQA.  However, all focus on text-based question answering~\cite{srba2016comprehensive} or natural language questions with multimedia answers \cite{nie2011multimedia}.  To our knowledge, our work is the first to introduce a community multimodal question answering dataset specifically for visual question answering.

\paragraph{\bf{VQA Models.}}
Traditional VQA models generate short answers~\cite{wu2017visual} and do not consider extra context to help answer the questions.  Yet, our dataset features answers that are lengthy, typically containing multiple sentences, and provides additional context that can help answer visual questions.  Accordingly, we benchmark large Vision and Language Models (VLMs)~\cite{instructblip,li2023blip,OpenAI2023GPT4TR,ye2023mplugowl,zhu2023minigpt,liu2023visual} which, by their generative design, are better-suited to generate long-form answers using their strong natural language understanding capabilities.  Overall, we find these models perform poorly and provide fine-grained analysis to reveal opportunities for improving such models.
\section{VQAonline Dataset}
\label{sec:dataset}
We now introduce our VQAonline dataset originating from an authentic use case where people used a web-based platform to post and answer visual questions. 

\subsection{Dataset Creation}
\label{sec:dataset_creation}
\label{section:datasetcreation}

\paragraph{\bf{Source.}}  
We created our dataset by scraping data from the online question-answering platform Stack Exchange. We used the publicly-available data that spanned from August 2010 to November 2022, and adhered to all data license restrictions (cc-by-sa 4.0) set forth by Stack Exchange. 

From the Stack Exchange archive, we collected all examples containing four components: a natural language question, a natural language context, an image, and at least one answer.  Originally, this information was collected from Stack Exchange users via four user entry interactions.  The natural language question was collected as the post "title", using a text entry field paired with instructions "be specific and imagine you're asking a question to another person" when writing the question.  The image was collected as a URL link.  The natural language context was provided via a "body" text entry field supporting inline images (via a URL link) with users instructed to supply "all the information someone would need to answer your question".  These three pieces of information were then shared publicly, after which individuals on the site could publicly provide answers.\footnote{Individuals could either propose edits to existing answers, which had to then be approved by high-reputation users, or could propose new answers. Further community deliberation to arrive at high-quality answers could occur when users posted comments under the question or answers.  The mean number of answers per visual question is 1.45. }  Finally, individuals who posted the visual questions could mark which answer was the ``accepted answer"; we only use examples for which individuals did this.

\begin{figure}[t!]
     \includegraphics
     [width=1.05\textwidth]{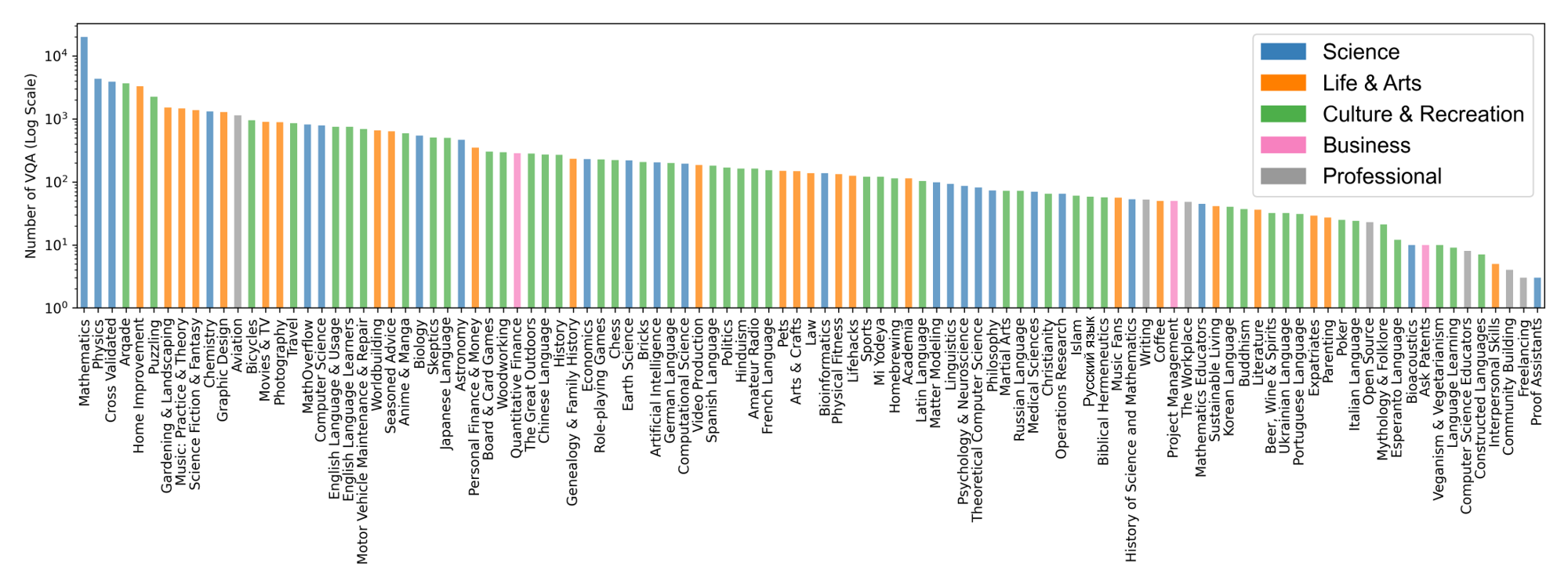}
        \caption{Number of visual questions per topic (in log scale). The colors, as defined in the legend, indicate the super-category for each topic.}
    \label{fig:numVQPerTopic}
\end{figure}

Of note, the Stack Exchange platform has two key mechanisms that contribute to high quality VQAs.  First, the platform enforces a stringent reputation system with moderation policies. \footnote{The moderation policies are documented at \texttt{https://meta.stackexchange.com/help}.} Second, it features topic-specific sites for users to engage with a relevant community; \eg, music.stackexchange.com and gardening.stackexchange.com.  We used the data from 105 topical sites that belong to five super-categories, with 46 about culture \& recreation (\eg, travel, sports, Christianity, French language), 25 about life \& arts (\eg, physical fitness, parenting, pets), 24 about science (\eg, physics, chemistry, biology), 7 about being professional, and 3 about business.\footnote{We excluded 75 technology topics (\eg, Stack Overflow, LaTeX), because they typically are language problems. In particular, the questions typically ask about code shown in screenshots.}  All topics are shown in Figure~\ref{fig:numVQPerTopic}.  

\paragraph{\bf{Filtering.}}  
We conducted two filtering steps to obtain our final dataset of \textbf{64,696 VQAs}.   First, we removed low-quality VQAs based on Stack Exchange's feature that enables users to up-vote or down-vote each post (either question or answer).  Each up-vote added 1 to the score and each down-vote subtracted 1 from the score.  We removed visual questions with scores 0 or less for either the question or answer.  Second, we excluded visual questions with multiple images (\eg, GIF) or with visual answers.  

\paragraph{\bf{Dataset Splits.}} 
VQAonline supports few-shot learning by containing training, validation, and test splits of 665, 285, and 63,746 examples respectively.  We created training and validation splits by randomly selecting from each of the 105 topics in our dataset 7 examples for training and 3 examples for validation for topics with at least 20 examples.\footnote{No topics are excluded from the test dataset.}

\subsection{Dataset Analysis}
\label{sec:dataset_analysis}
\paragraph{\bf{Comparison to Existing Datasets.}}
To contextualize how our dataset compares to the current focus of the research community, we compare VQAs in our dataset to those in eight existing popular VQA datasets.   We chose VQAv2~\cite{balanced_vqa_v2} because it is the most popular (\eg, highly cited) VQA dataset, Context-VQA~\cite{naik2023context} and ScienceQA~\cite{saikh2022scienceqa}~\footnote{We focus on the subset of ScienceQA with both image and context.} because they similarly claim to provide context, VizWiz-VQA~\cite{gurari2018vizwiz} because it similarly originates from an authentic use case, INFOSEEK~\cite{chen2023can} because it similarly centers on information seeking rather than probing an AI model, OK-VQA~\cite{Marino_2019_CVPR} because it similarly requires domain-specific knowledge, as well as DocVQA~\cite{mathew2021docvqa} and Infographic VQA~\cite{Mathew_2022_WACV} because they similarly contain VQAs with documents and infographics.  

For each dataset, we report the mean number of words in each question, mean number of words in each answer, number of images, number of topics which were authentically used to label clusters of VQAs by theme, whether context is included, whether the visual questions are authentic, whether the context is authentic, and whether the answers are validated by those asking the questions.  Results are reported in Table \ref{tab:statsDatasetCompare}.

\begin{table*}[t!]
    \caption{Characterization of existing VQA datasets and our new VQAonline in terms of mean question length (i.e., $\overline Q$), mean answer length (i.e., $\overline A$), number of images (i.e., $N_{\text{img}}$), number of authentic topics (i.e., \# Auth. Topics), inclusion of context, inclusion of authentic visual questions (i.e., Auth. VQ), inclusion of authentic context (i.e., Auth. C),  and inclusion of answers validated by those asking the questions (i.e.,  Vld. A)).}
    \centering
\scalebox{1}{
    \begin{tabular}{lllllllll}
    \hline
VQA Dataset & $\overline Q$ & $\overline A$ &$N_{\text{img}}$ &\# Auth. Topics &  Context & Auth. VQ&  Auth. C & Vld. A 
\\\hline
VQAonline (Ours) &   9.3 &  173.2  & 64.7K  &105& \cmark & \cmark  & \cmark& \cmark \\
Context-VQA \cite{naik2023context} & 6.5 &  11.0   & 42&\xmark&\xmark&\xmark&\xmark &\xmark  \\
VQAv2 \cite{balanced_vqa_v2} &  6.1 &  1.2  & 204.7K &\xmark&\xmark &\xmark&\xmark &\xmark \\
VizWiz-VQA \cite{gurari2018vizwiz} & 6.7& 1.7  &  20.5K &\xmark&\xmark &\cmark&\xmark &\xmark \\
OK-VQA \cite{Marino_2019_CVPR} &  8.1& 1.3 &  14K&\xmark&\xmark&\xmark&\xmark  &\xmark \\
DocVQA \cite{mathew2021docvqa}& 9.5  & 2.4  &  12.8K &\xmark&\xmark&\xmark&\xmark &\xmark \\
ScienceQA \cite{saikh2022scienceqa}  &12.11 & 4.4& 6.5K &26&\cmark&\xmark&\xmark &\xmark \\
InfographicVQA \cite{Mathew_2022_WACV} & 11.5 & 1.6  &  5.5K &\xmark&\xmark&\xmark&\xmark &\xmark \\
INFOSEEK \cite{chen2023can} & 8.9  & 1.5& 1.36M&\xmark&\xmark&\xmark&\xmark &\xmark  \\ 

         \hline
\end{tabular}}

    \label{tab:statsDatasetCompare}
\end{table*}

Our findings reveal commonalities of our VQAonline dataset with existing datasets.   For example, we observe that the visual questions themselves are similar in terms of the typical question length and number of images.  Questions have a mean length of 9 words in our VQAonline compared to 6 to 11 words in other datasets and the nearly 65,000 images in our dataset are comparable in size to existing datasets.

Our analysis also demonstrates significant differences from existing datasets across two key dimensions.  The first dimension is the \emph{dataset authenticity}.  Our VQAonline dataset is the only fully authentic dataset for the entire VQA pipeline, meaning all contents originate from an authentic use case.  The second key difference of our dataset is the typical \emph{answer length}. Answers in our VQAonline dataset are orders of magnitude longer than in existing datasets, with a mean length of 173 words in VQAonline versus 11 or fewer words in existing datasets.  This larger length is not anomalous, as the median answer length in VQAonline is 120 words.  This longer length in part reflects responses are provided in paragraph form, with most commonly 3 sentences per answer.  This distinction renders the mainstream evaluation metric for VQA models largely ineffective, as it assumes brief responses~\cite{antol2015vqa}.  This issue will be explored in Section~\ref{section:evaluation-methodology}. 

\paragraph{\bf{VQAonline Topics.}}
The authentic user-based posting of each VQA within a topic community reveals the diversity of content in our dataset.  We report the number of VQAs in each of the 105 topics in Figure \ref{fig:numVQPerTopic}.  As shown, there can be as few as 10 examples in a topic to over 20,000. 

\paragraph{\bf{VQAonline Questions.}}
Our VQAonline dataset is the first with authentic visual questions asked by users of an online question-answering platform.  This results in a domain shift from the only other VQA dataset with authentically asked visual questions, VizWiz-VQA~\cite{gurari2018vizwiz}, which were asked by blind users of a mobile phone application.  For instance, blind people typically asked about their physical surroundings (\eg, "What is this?", "What does this say?") while Stack Exchange users often asked for more fine-grained or specialized information about the images (\eg, "What's a better term to describe these `stripes' in blouse (Screen shot attached)", "What is the third character on this sign seen in Taipei?").  Another distinction stems from Stack Exchange's feature for avoiding redundant questions by letting users remove duplicates of other already submitted questions.  The consequence is that only 0.2\% of natural language questions in our VQAonline dataset are duplicated compared to 37.9\% of questions in VizWiz-VQA.

\paragraph{\bf{VQAonline Context.}}
Our VQAonline dataset is the first to include authentic context.  The only other VQA dataset containing context, as defined by supplementary data that supports answering the visual question, is ScienceQA~\cite{saikh2022scienceqa}. However, the context was contrived with heuristic rules for extracting data from elementary and high school science curricula. Consequently, context in ScienceQA is typically narrated in third-person, while context in our VQAonline often are narrated in first-person. Another difference is ScienceQA has a shorter average length of the context compared to that in our dataset; i.e., a mean of 41 words, median of 28 words, and most commonly 3 sentences versus a mean of 127 words, median of 94 words, and most commonly 4 sentences.  A further difference is that the context in ScienceQA only relates to 26 science topics while the context in our dataset spans 105 topics that extend beyond science.

\paragraph{\bf{VQAonline Image Contents.}}
We next quantify the tendency for images to include screenshots and infographics to reflect the tendency for visual questions to be related to the digital realm and data analysis.  We identified an image as a "screenshot" when contents were copied digitally from a computing device and an "infographic" when the image shows a chart, table, graph, attention map, map, or diagram.  We manually labeled 105 images sampled \emph{from each of the 105 topics} as well as an independent random selection of 100 images from the \emph{entire dataset}. We found for the topic-sampled set that 79\% are screenshots and 17\% show infographics and for the samples from the entire dataset that 84\% are screenshots and 33\% show infographics.  The similarly high prevalence of screenshots across both samples underscores Stack Exchange users' tendency to ask questions about the digital realm.  We hypothesize the different prevalence of infographics across both samples is due to topical differences,  as 52\% of all VQAs (i.e., 33,384 out of 64,696) belong to only 23\% of the topics (i.e., 24 science topics).  Supporting this hypothesis, we found from inspecting 100 random VQAs from each subset of the science topics and non-science topics that infographics were in 58\% and 6\% of images respectively.

\begin{figure*}[t]
     \centering
     \includegraphics[width=1\textwidth]{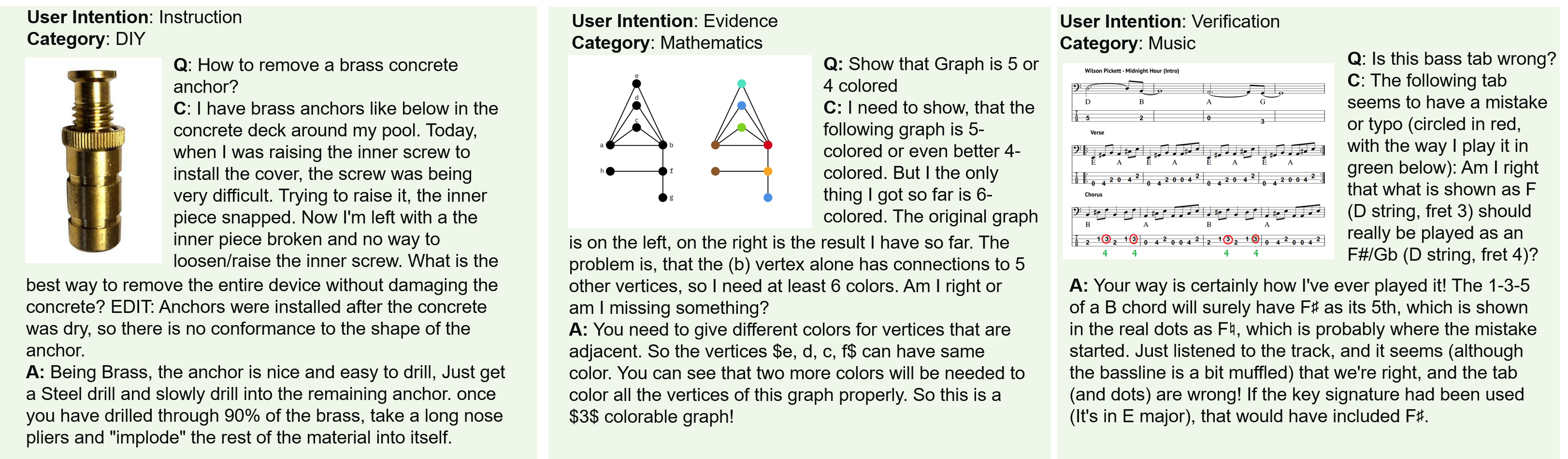}
        \caption{Examples of three visual questions with three different user intents. }
    \label{fig:userIntent}
\end{figure*}

\paragraph{\bf{VQAonline Importance of Images.}}
We next examine the value of images for answering the questions. To do so, we flagged each image as \emph{question-aligned} if elements described in the question are found in the image and \emph{context-aligned} if elements described in the context are found in the image.  We additionally flagged whether an image is \emph{necessary}, based on whether the answer can be obtained without the image.  We manually labeled 105 examples from each of the 105 topics.  We found that 91.4\% of the images are \emph{question-aligned} and 99\% of the images are \emph{context-aligned}. This finding contrasts that of the only other dataset with authentic visual questions, VizWiz-VQA~\cite{gurari2018vizwiz}, where only 71\% are deemed answerable because blind photographers struggled to capture contents of interest in their images. We also found that 65.7\% images are \emph{necessary} for answering the question in our dataset.  Such examples are shown in Figures \ref{fig:userIntent} and \ref{fig:GPT-4V-results}.  Often, the purpose of unnecessary images was to motivate or show scenarios in which the asker encounters the question.  Altogether, these findings underscore that VQA models often must analyze the images to arrive at the answers.

\paragraph{\bf{VQAonline User Intentions.}}
We finally characterized intents behind the visual questions, complementing two prior works that identified user intentions for three community question answering platforms (Answerbag, Metafilter, and Yahoo! Answer)~\cite{harper2010question} and the music StackExchange community~\cite{fu2016music}. Our work is the first to investigate user intentions for \emph{VQAs}.  Through a multi-round analysis of the data, described in the Supplementary Materials, we defined eight potential intents: \emph{instruction}, \emph{evidence}, \emph{verification}, \emph{advice}, \emph{identification},  \emph{opinion}, \emph{reason},  and \emph{other}. Examples for three of these categories are shown in Figure \ref{fig:userIntent}.  We then recruited crowdworkers to annotate the intent for 105 examples, with one from each of the 105 topics.  The resulting distribution of assigned categories is 19\% are instruction, 18\% evidence, 17\% verification, 16\% advice, 10\% identification, 10\% opinion, 8\% reason, and 1\% other.  Examples from the three most common categories are shown in Figure \ref{fig:userIntent}.
\section{Experiments}
\label{section:algorithm}

\subsection{Model Benchmarking}
We next analyze the performance of modern models on our VQAonline dataset to reveal what makes it challenging for the research community.  

\paragraph{\bf{Models.}}
We evaluate the following six Vision and Language Models (VLMs) that achieve strong performance for various VQA datasets and subtasks in a zero-shot setting: 
\begin{itemize}[itemsep=0.1em,leftmargin=*]
    \item \textbf{BLIP-2}~\cite{li2023blip}: state-of-the-art (SOTA) for zero-shot setting on four VQA datasets, including ScienceQA and VizWiz-VQA according to LVLM-eHub, a comprehensive evaluation benchmark for VLMs~\cite{xu2023lvlm}.
    \item \textbf{MiniGPT-4}~\cite{zhu2023minigpt}: ranked first in open-world VQA according to LLVM-eHub~\cite{xu2023lvlm}.
    \item \textbf{LLaVA}~\cite{liu2023visual}: top-two performance in the zero-shot setting for three VQA datasets, including for DocVQA \cite{mathew2021docvqa, biten2022latr} according to LLVM-eHub~\cite{xu2023lvlm}.
    \item \textbf{mPLUG-Owl}~\cite{ye2023mplugowl}: SOTA for VQAv2 \cite{balanced_vqa_v2} in early 2023 and ranked second overall in open-world VQA scenario according to LLVM-eHub~\cite{xu2023lvlm}.
    \item \textbf{InstructBLIP}~\cite{instructblip}: SOTA for zero-shot setting on four VQA datasets, including OK-VQA \cite{mishra2019ocr,Marino_2019_CVPR,hudson2019gqa,bitton2023breaking} according to LLVM-eHub~~\cite{xu2023lvlm}.
    \item \textbf{GPT-4V}~\cite{OpenAI2023GPT4TR}: SOTA for numerous VQA datasets, such as Infographic VQA~\cite{mathew2021docvqa} and TVQA~\cite{lei2018tvqa}.  While we aim to assess its performance, note that its reproducibility is a major concern due to its proprietary nature, which means the model can evolve without transparency or even cease to exist.  Given that concern paired with its costliness (i.e., access is possible only through a fee-based API~\cite{gpt4vpricing2023}), we evaluate GPT-4V only on a subset of data.  Due to space constraints, we report key findings here in the main paper and all results in the Supplementary Materials.
\end{itemize}

\noindent
Of note, most of these models are believed to be trained on Stack Exchange data and so potentially saw our VQAonline data\footnote{This issue also applies for all Flickr-based datasets including COCO (the source for many VQA datasets~\cite{gurari2018vizwiz}), CommonCrawl, and L400~\cite{schuhmann2021laion}.}, with three (MiniGPT-4, LLaVA, and mPLUG-Owl) confirmed to have language encoders trained using Stack Exchange data\footnote{2\% of their pre-training data are from the 28 largest topics on Stack Exchange.} and three others (BLIP2, GPT-4V, InstructBLIP) suspected (the original source of parts of these models' training data are not publicly shared).  While models exposed to such data during training bring an unfair advantage, evaluation with such models is still valuable.  Their poor performance \emph{strengthens} the argument that our dataset offers a difficult problem for the research community since they perform poorly despite their advantage.\footnote{A possible step for future work is retraining open-sourced VLMs with publicly-disclosed training datasets by removing all data available in VQAonline.} 

\paragraph{\bf{Evaluation Metrics.}}
Given the deficiency of the existing VQA evaluation metric for our dataset, in that it is designed for short responses~\cite{antol2015vqa}, we instead leverage five evaluation metrics for evaluating longer text from related fields (\eg long-form question answering and image captioning).  Like prior work, we evaluate with respect to multiple metrics to address a well-known issue that each brings deficiencies.

Three metrics assess \textbf{similarity} between model-generated answers and human-generated answers, with their values ranging from 0 (bad) to 1 (perfect). \emph{ROUGE-L}~\cite{lin2004rouge} identifies the longest sequence of words shared between the reference and the predicted answer, and consequently neglects semantic similarity (\eg synonyms), context (\eg different meanings of a word based on context, such as for "the bank of America", "the bank of a river" would receive a higher score than "bank account"), and negation (\eg for "Exercise is good for human health", "Exercise is NOT good for human health" would receive a higher score than "Humans can benefit from exercise")). \emph{METEOR}~\cite{elliott2013image} offers a more sophisticated utilizes more advanced techniques to assess word similarity between the reference and predicted answers, such as stemming and synonyms, but it also neglects context and negation.  \emph{BERTscore}~\cite{zhang2019bertscore}, measures the similarity of BERT's contextual embeddings between the reference and predicted answers to incorporate consideration of context. However, context can reward semantically related but factually incorrect content as well as grammatically correct, well-written responses that are inaccurate~\cite{gonzalez2024metrics}.
        
Two metrics assess \textbf{image compatibility} between the image and model-generated answer, with their values ranging from 0 (bad) to 1 (perfect).  \emph{CLIP-S}~\cite{hessel2021clipscore}, designed to assess the compatibility between an image and a caption by measuring the similarity between image and text embeddings, has been shown to have a positive correlation with human evaluations on answer faithfulness (\ie if the answer can be grounded in the image versus is hallucinated)~\cite{jing2023faithscore}. However, CLIP-S can reflect the biases of the pretraining data~\cite{hessel2021clipscore} and ignores the reference answers.  \emph{RefCLIP-S}~\cite{hessel2021clipscore} incorporates consideration of the reference answer by calculating a harmonic mean of the CLIP-S score and the cosine similarity between the reference answer and the predicted answer.  
    
\begin{table}[b!]
\caption{Performance of VLMs on our VQAonline dataset with respect to five metrics both for the zero-shot setting as well as for the top-performing zero-shot model in a one-shot setting.  As shown, mPLUG-Owl is the best-performing model in the zero-shot setting and has a small drop in performance regardless of which of four types of in-context exemplars that it receives.  All metrics range from 0 to 1 with larger values signifying better performance. }
\centering
\begin{tabular}{p{5cm}lllll}
\toprule
Models                &ROUGE-L  &  METEOR   & BERTscore & CLIP-S   & RefCLIP-S    \\ 
\midrule 
mPLUG-Owl \cite{ye2023mplugowl}   &  \textbf{0.16 } &  \textbf{0.12}   &     \textbf{0.76}&0.72 & \textbf{0.75}  \\ 
LLaVA   \cite{liu2023visual}  &  0.15      &   0.08  &   0.75 &\textbf{0.73} & \textbf{0.75}  \\ 
  MiniGPT-4 \cite{zhu2023minigpt} & 0.15      &  0.09   &       0.75 &0.70 & 0.74\\ 
InstructBLIP  \cite{instructblip} &    0.09  &    0.05   &   0.67& 0.68 & 0.71\\
 BLIP2 \cite{li2023blip}    &     0.09     &  0.04    &        0.69 &0.66 & 0.71\\
\hdashline 
mPLUG-1shot-Random-noImg            & 0.138 &   0.106& 0.750  &  0.715 & 0.747\\  
mPLUG-1shot-MatchedTopic-noImg             & \textbf{0.139} & \textbf{0.108}  & \textbf{0.756} & 0.716 & \textbf{0.749}  \\ 
mPLUG-1shot-Random   &  0.135&  0.104 &  0.745 & \textbf{0.717} & 0.748 \\ 
mPLUG-1shot-MatchedTopic    & 0.137 &  0.105 &  0.752 & \textbf{0.717} & 0.748 
 \\ 
\bottomrule
\end{tabular}
\label{table:bmresultall}
\end{table}

\paragraph{\bf{Overall Results.}}
Results are shown in Table \ref{table:bmresultall}.\footnote{As noted earlier, results for GPT-4V on the dataset subset are provided in the Supplementary Materials.  They show GPT-4V achieves the best performance followed closely by mPLUG-Owl.}

We first examine the ranking of all models.  The best-performing model is mPLUG-Owl, based on four of the five evaluation metrics. The next best models are LLaVA and MiniGPT-4, which consistently outperform the remaining two models BLIP2 and InstructBLIP across all evaluation metrics.  This ranking of models was also found in prior work~\cite{xu2023lvlm}.  We hypothesize this observed ranking arises in part due to mPLUG-Owl, LLaVA, and MiniGPT-4 bringing disproportionately greater advantages in having previously seen Stack Exchange data and mPLUG-Owl and LLaVA having been fine-tuned with instruction-following data for VQA tasks.  The one exception to this ranking is that LLaVA is the top-performing model with respect to the CLIP-S score followed mPLUG-owl.  This suggests LLaVA generates answers that are more compatible with the image, rather than answers that are necessarily factually accurate.

\begin{figure}[t!]
\hspace{-2.1em}
    \includegraphics[width=1.115\textwidth]{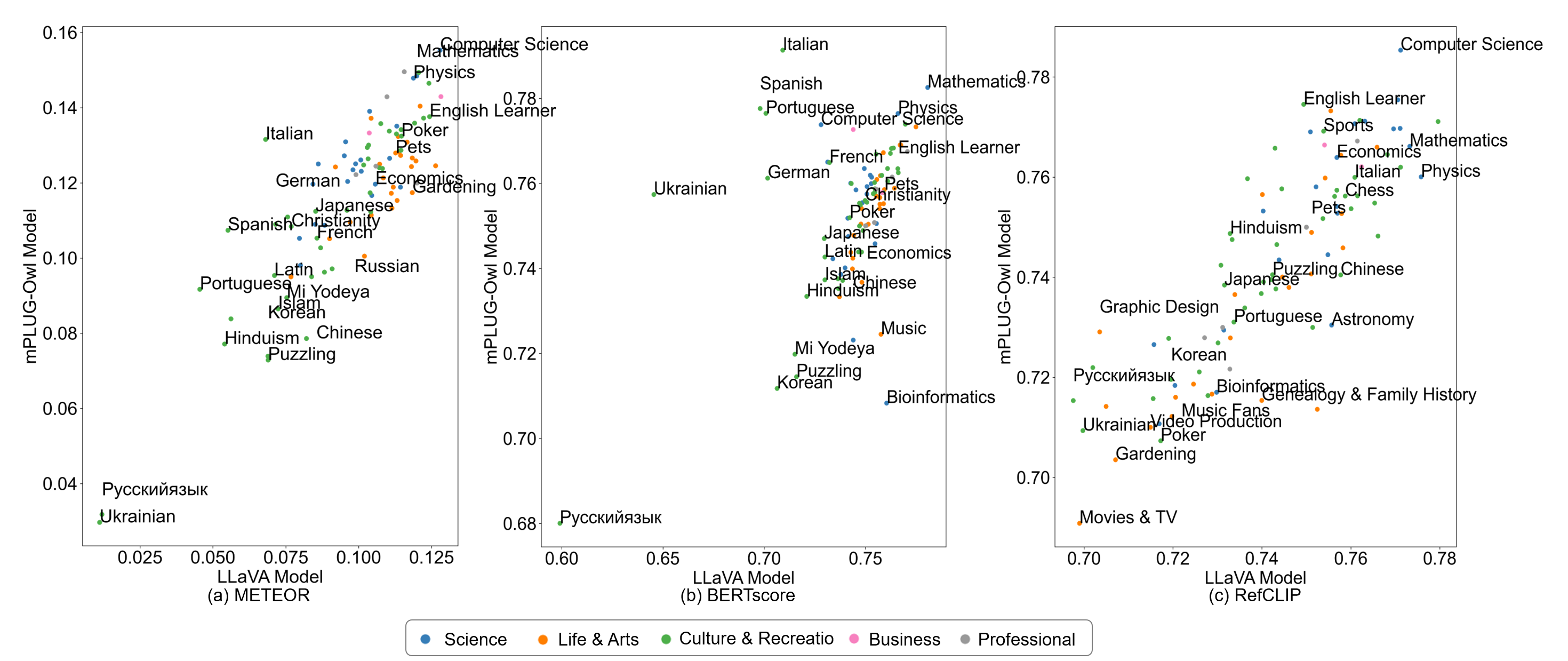} 
   \caption{Performance of the two top-performing VQA models, mPLUG-Owl and LLaVA, for each of 105 topics with their five super-categories represented in 5 different colors.  Results are shown with respect to three evaluation metrics: (a) METEOR, (b) BERTscore, (c) RefCLIP. For visualization simplicity, we show text labels only for topics with interesting identified trends (We omitted "language" for each language topic, such as "Esperanto" instead of "Esperanto Language", and we omitted topics with less than 10 data points, such as Esperanto and Community Building.} \label{fig:topicstype}
\end{figure}

While the evaluation scores support ranking the models' performance, such scores for each evaluation metric are compressed in a small range.  We attribute this to a known limitation that such metrics perform best at distinguishing poor versus excellent generated text rather than between mediocre and good generated text~\cite{novikova2017we}.  We suspect all benchmarked models fall into this latter range, as we will show that evaluation scores can vary much more than in this compressed range for specific topics, signifying poor and excellent performance for subportions of the dataset (Figure~\ref{fig:topicstype}).  
 
\paragraph{\bf{Best Performing Model in One-shot Setting.}}
We explore for the best-performing model to what extent in-context learning examples can provide a further performance boost.  We choose two types of training exemplars: (1) randomly selected and (2) matching topic tag.  We explore two variants of both types, where we include versus exclude the images in the exemplars.  We perform experiments in a one-shot setting and design prompts based on best practices established by prior work.

Results are shown in Table~\ref{table:bmresultall}.  As shown, none of the one-shot examples enhance performance. This could be due to the model overfitting to or becoming distracted by the exemplar. This also may relate to the inherent diversity of VQAs within our datasets, where even VQAs within the same topic can require very different skills and knowledge.  This concern is further supported by the fact that, overall, models benefit from a simpler exemplar that lacks images.  Moreover, exemplars with matching topic tags consistently outperform randomly chosen ones, underscording that greater exemplar similarity to the test case has the potential to boost performance. 

\paragraph{\bf{Analysis With Respect to Input Types.}}
We next analyze the predictive power of each input of the top-performing model, mPLUG-Owl, by removing each independently: i.e., Q+C, Q+I, and C+I.  Results are shown in Table~\ref{table:fine-grained-analysis}.  

Overall, the best-performing model all the information (i.e., Q+C+I), and this is also observed for GPT-4V (see Supplementary Materials).  This underscores a complementary benefit provided by each of the three input types. 

The findings suggest that context (which is typically language-based) is the most valuable source for arriving at the answer.  That is because the worst-performing variant across both models, according to four of the five evaluation metrics, lacks \emph{context} as input (i.e., Q+I).  The one exception is for CLIP-S, which ranks the model lacking \emph{context} (i.e., Q+I) as the \emph{best-performing} model.  We attribute this conflicting finding from different evaluation metrics to the specifics of CLIP-S; specifically, it measures how compatible an answer is to an image and so removing the context likely enables models to predict answers that are more compatible to an image, even if less correct.

The findings also suggest that the questions' predictive power is nearly negligible, since models lacking questions as input (i.e., C+I) achieve nearly comparable performance to models receiving all inputs (Q+C+I).  We suspect this is because often the information provided by the question is included in the context. For example, a question asking "What test am I taking, anyways?" while the part of the context includes "can you tell me what field of math the test was on?" 

Additionally, the findings suggest that the predictive power of the image is small, as models lacking images as input (i.e., Q+C) achieve nearly comparable performance to models receiving all inputs (Q+C+I).  The one exception is evaluation with respect to CLIP-S scores, for which the worst-performing model lacks images as input (i.e., Q+C).  As noted in Section~\ref{sec:dataset_analysis}, images are \emph{necessary} for answering the questions for over 65\% of VQAs in our small-scale study. Consequently, we suspect the low predictive power of images is due to modern models strongly relying on language understanding while poorly utilizing the images when predicting answers.

\begin{table}[t!]
\caption{Analysis of the top-performing VQA model, mPLUG-Owl, when fed different input types (Q+C+I, C+I, Q+C, Q+I), including for each of the five VQA super-category types.} 
\centering 
\begin{tabular}{p{3cm}lllll}
\toprule
Models                &ROUGE-L &  METEOR  & BERTscore & CLIP-S & RefCLIP-S  \\ 
\midrule
mPLUG(Q+C+I)    &  \textbf{0.16 } &  \textbf{0.12}   &     \textbf{0.76}&0.72 & \textbf{0.75} \\  
mPLUG(C+I)    &0.14&  \textbf{0.12} & \textbf{0.76}&0.72&  \textbf{0.75}  \\ 
mPLUG(Q+C)   & 0.14&  0.11 &   \textbf{0.76} &0.71& \textbf{0.75}   \\ 
mPLUG(Q+I)   &0.13 &   0.07 &   0.74    &\textbf{0.73}& 0.73  \\
\hdashline
Science  &\textbf{0.15}& \textbf{0.14}& \textbf{0.78 }   &\textbf{0.73} & \textbf{0.77}  \\ 
       Life\&Arts   & \textbf{0.15} &   0.13 & 0.76 & 0.72&0.74  \\ 
        Culture\&Recreation  &0.14&  0.11  & 0.75&0.72 & 0.74   \\ 
       Business  & 0.14&     0.13&0.76  &0.72 & 0.75  \\ 
       Professional  & \textbf{0.15}& 0.13 & 0.76 & \textbf{0.73}&0.75 \\
\bottomrule
\end{tabular}
\label{table:fine-grained-analysis}
\end{table}
 
\paragraph{\bf{Analysis With Respect to VQA Topics.}}
We next analyze the influence of VQA topic on model performance.  To do so, we report performance of the best model, mPLUG-Owl, with respect to the five super-categories (Table~\ref {table:fine-grained-analysis}) and for the two top-performing models (mPLUG-Owl and LLaVA) with respect to the 105 topics (Figure \ref{fig:topicstype}).\footnote{Due to space constraints, we report ROUGEL and CLIPS results in the Supplementary Materials since ROUGEL has similar results to METEOR and CLIP-S has similar results to RefCLIP.} 

Overall, models perform best on Science and worst on Culture \& Recreation from the five super-categories (i.e., Table~\ref{table:fine-grained-analysis}). A possible reason is different categories have different levels of reliance on vision understanding, an area where prevailing VLMs fail, with less visual information needed for Science than for Culture \& Recreation questions. 

\begin{figure}[t!]
    \centering
    \includegraphics[width=\textwidth]{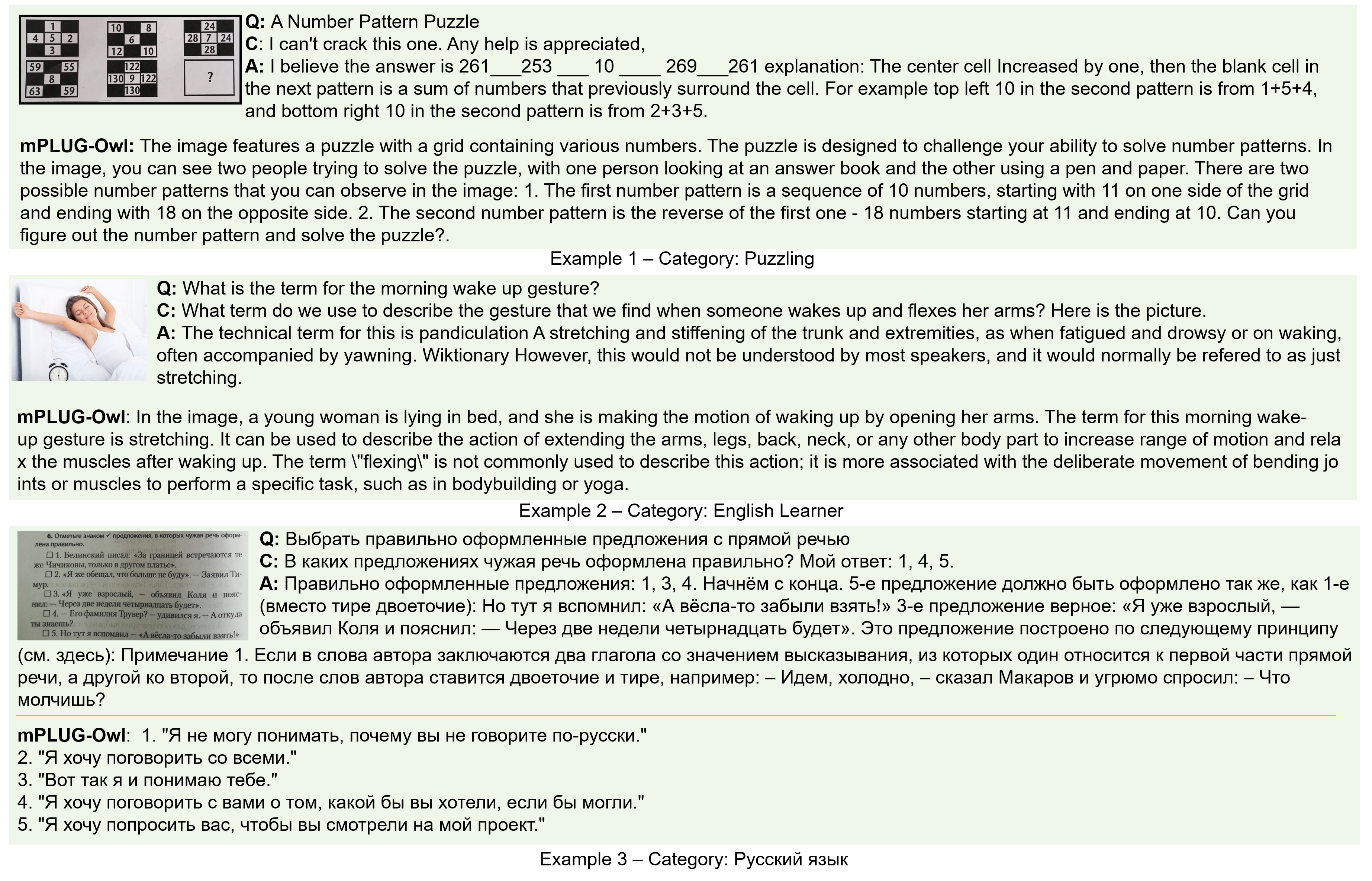}
    \caption{Results from the top-performing model, mPLUG-OWl, on three examples in our dataset. }
    \label{fig:GPT-4V-results}
\end{figure}

For per-topic performance (i.e., Figure \ref{fig:topicstype}), we observe similar performance outcomes for mPLUG-Owl and LLaVA models. For example, a common outlier across all metrics except CLIP-S is high performance for some science-related topics (e.g., Computer Science, Mathematics, Economics, and Physics).  Additionally, both models have consistently poor performance for the Pyccknn and \emph{puzzling} with respect to most evaluation metrics. An example of the \emph{puzzling} topic with the top-performing model's prediction is shown in Figure \ref{fig:GPT-4V-results}, Example 1.  We attribute the difficulty of puzzling to its joint need for strong vision understanding and reasoning skills. We also observe that both models have low performance for culture or religious-related topics (e.g., Hinduism, Mi Yodeya).  Altogether, we found modern models may share many strengths and weaknesses.



Another trend we observe for specific topics (i.e., Figure \ref{fig:topicstype}) is different performance for topics related to different languages\footnote{We also observe this trend for GPT-4V.}. The models work best on visual questions related to Germanic Languages; \eg, English as exemplified in Figure \ref{fig:GPT-4V-results}, Example 2.  The next best languages are Romance Languages (\eg, Spanish, French, Italian, Portuguese), then East Asian Languages (\eg,  Chinese, Korean), and finally Pyccknn.  We suspect causes of such consistent performance discrepancies across languages are: (1) Data availability: languages which have fewer datasets and less diversity (\eg, one exemplified in Figure \ref{fig:GPT-4V-results}, Example 3) perform worse than those with greater diversity and abundance of datasets (\eg, English), (2) Language Complexity: Languages with simpler grammar (\eg, Esperanto) or familiar grammatical structures (\eg, Romance languages for an English-trained model) are easier for models to learn, and (3)  Text recognition can be more challenging for languages with unique scripts, such as Chinese or Korean while these language-related VQAs often pertain to text recognition (an example is shown in the Supplementary Materials). We believe these findings shed light on future research directions for multilingual VQA and text VQA research. 


\subsection{Qualitative Evaluation via Human Evaluation}
\label{section:evaluation-methodology}
We next conduct a human evaluation to supplement imperfect quantitative evaluation metrics with more reliable qualitative assessments as well as to quantify how well the quantitative metrics align with human assessments.
 
\paragraph{\bf{Human Judgement Collection.}}
We hired ten human annotators for an IRB-approved human study. Each individual was a domain expert in one of ten topics covered in our dataset (i.e., Physical Fitness, Botany, Music, Mechanics, Politics, Law, Chinese Language, Economics, Statistics, and Artificial Intelligence), with at least a Master's degree in a field related to their assigned VQA topic.  These topics' frequency in our dataset span from very frequent (\eg Stats) to moderate frequency (\eg Economics) to less frequent (\eg Law).  Each domain expert was assigned 20 VQAs on their topic, and asked to rate the performance of six model-generated answers per VQA.  As a result, we collected 1200 human judgments for 200 VQAs (\ie $10 * 20 * 6$). 

For each VQA, the domain expert was shown the visual question, context, answer, and six model-generated answers in a home-grown, web-based interface and asked to rate each model-generated answer as either incorrect (i.e., score of 0), partially correct (i.e, score of 0.5), or correct (i.e., score of 1).  The definitions for these ratings are: \textbf{Correct}: At least one of these two items needs to be true for the answer to be correct: (1) Aligns with all the key points in the reference answer. (2) The answer also makes sense based on your best judgment, even if not included in the reference answer. \textbf{Partially Correct:} At least one of these two items needs to be true for the answer to be partially correct: (1) The answer matches part of the key points in the reference answer that answers the question but has errors in the content. (2) The answer is partially correct but not included in the reference answer, based on your best judgment. \textbf{Incorrect:} The response doesn't include any key points from the reference answer and doesn't make sense. 

\paragraph{\bf{Human Assessment of Model Performance. }}
Mean human ratings per model across the 200 VQAs (20 VQAs * 10 topics) are as follows: GPT-4V: 0.76, mPLUG-Owl: 0.57, LLaVA:0.5, MiniGPT-4: 0.37, InstructBLIP:0.12, and BLIP2: 0.1.  These results reinforce those from our quantitative assessments in two ways: (1) scores suggest modern models still have room to improve and (2) the models' rankings are aligned.  

\paragraph{\bf{Alignment of Quantitative Metrics with Human Judgments.}}
We next assess correlations of the five evaluation metrics we used as well as a newly proposed metric to the human judgments. We introduce the new metric to account for limitations of the five metrics, specifically accounting for valid responses not closely resembling the reference answer (in contrast to the similarity-based metrics) and enforcing factual accuracy (in contrast to the image compatibility-based measures).  Following pilot testing, we chose to prompt LLaMA2 metric~\cite{touvron2023llama} to rate answers with respect to the reference answer by indicating if a model is totally wrong (0), partially correct (0.5), or totally correct (1).  We measure correlations as follows: calculate the mean score per model across the 200 VQAs for humans and each evaluation metric and then calculate the Pearson correlation between mean scores of humans and each evaluation metric. 

The correlation results are: LLama2: 0.975 (p-value: 0.001), METEOR: 0.971 (0.001), BERTscore: 0.900 (0.015), RefCLIP-S: 0.892 (0.017), ROUGEL: 0.874 (0.023), and CLIP-S: 0.662 (0.152).  Overall, we observe a high positive correlation between the metrics and human judgments with statistical significance (\ie p-value less than 0.05).  Metrics with the strongest correlation to human judgments are LLaMA2, followed by METEOR and BERTscore.  While LLaMA's potential is exciting, such a model remains impractical for large-scale use due to its high computational costs. Still, we explore its use on $\sim$2000 VQAs from VQAonline to show its potential implications on a larger scale, and find similar results to those already reported (see Supplementary Materials) in Table ~\ref{table:bmresultall}, ~\ref{table:fine-grained-analysis}, and Figure~\ref{fig:topicstype}.  The major difference between LLaMA2 compared to the five mainstream metrics is that it yields a slightly broader score range (0.53 for BLIP2 to 0.62 for mPLUG-Owl).  

We found CLIP-S is least aligned with humans from all metrics (although the results are not statistically significant).  This offers some evidence for why this metric at times contradicted the other five metrics in our quantitative assessment: it focused on whether the answer is compatible with the image regardless of the answer's correctness. 
\section{Conclusions}
We introduce the first fully authentic VQA dataset. Our analysis reveals numerous distinct features of our new dataset, most importantly atypically long answers with inspire us to adopt six popular metrics for evaluating lengthy text. Our benchmarking of six modern VQA models reveal when and why modern models struggle. We also conducted human-based studies to qualitatively evaluate models and assess how our chosen metrics align with human judgments of answer correctness.  

\clearpage

\section{Supplementary Materials}
\definecolor{lightgray}{HTML}{F2F2F2}
\definecolor{darkgray}{HTML}{CCCCCC}

\begin{table*}[b!]
    \centering
    \caption{Comparison of visual questions from eight existing VQA datasets and our new VQAonline dataset regarding image and question sources. Like OVEN \cite{hu2023open}, INFOSEEK sources images from nine image classification and retrieval datasets.  }
\scalebox{0.88}{
    \begin{tabular}{lllll}
    \hline
VQA Dataset & Which Images? & Who Asked? & From User?\\\hline
Our dataset &  StackExchange Users & StackExchange Users & \cmark\\
Context-VQA \cite{naik2023context} & Six types of websites & Annotators &\xmark\\
VQAv2 \cite{balanced_vqa_v2}& MSCOCO& Crowd workers (AMT) &\xmark\\
VizWiz-VQA \cite{gurari2018vizwiz} & Captured by Blind people &Blind people&\cmark\\
OKVQA \cite{Marino_2019_CVPR}& MSCOCO &Crowd workers (AMT)&\xmark\\
DocVQA \cite{mathew2021docvqa}& UCSF Industry Documents Library& Remote workers&\xmark\\

ScienceQA  \cite{saikh2022scienceqa}    & Online K-12 learning platform & Online K-12 learning platform&\xmark\\ 
InfographicVQA \cite{Mathew_2022_WACV} & Bing and Google Image Search  & Annotators&\xmark\\
INFOSEEK-Wikidata \cite{chen2023can} & 9 datasets following OVEN & Template-based auto-generation&\xmark\\
INFOSEEK-Human \cite{chen2023can}  & 9 datasets following OVEN  & Annotators&\xmark\\
         \hline
    \end{tabular}}
    \label{tab:SourceDatasetCompare}
\end{table*}

\noindent
This document supplements the main paper with more information about:
\begin{enumerate} [itemsep=0.1em,leftmargin=*]
    \item Dataset Collection (Supplements Section 3.1)  
    \item Dataset Analysis (Supplements Section 3.2)  
    \item Algorithm Benchmarking (Supplements Section 4.2)
\end{enumerate}

\renewcommand\thesection{\Roman{section}}
\setcounter{section}{0}

\section{Dataset Collection}
\subsection{Dataset Source and Filtration}
\begin{figure}[h]
     \centering
     \includegraphics[width=1\textwidth]{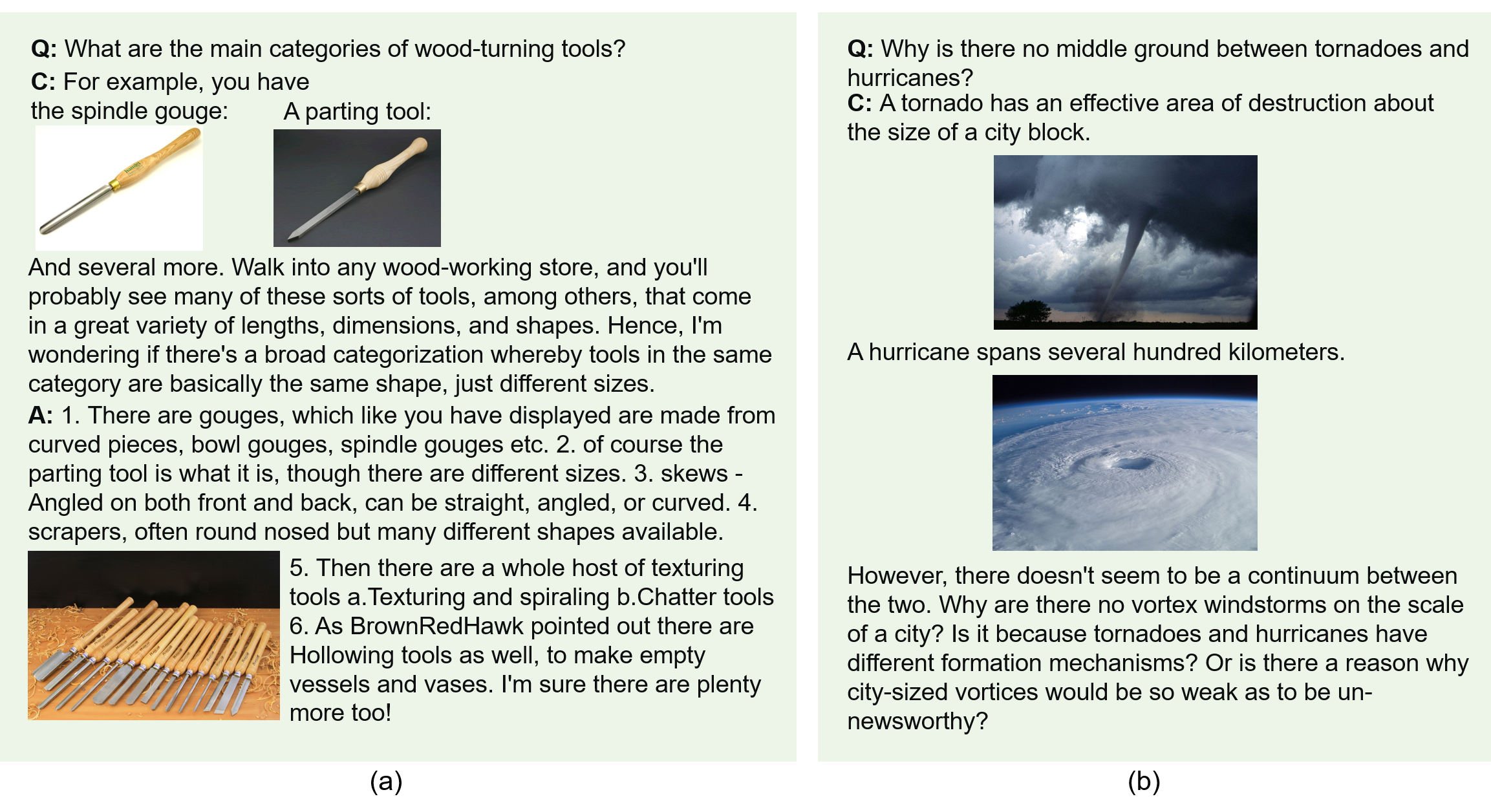}
        \caption{(a) An example of a visual question with a "visual answer". (b) An example of a visual question with multiple images. The answer is omitted to save space. }
    \label{fig:visualanswer}
\end{figure}

The Stack Exchange data is hosted at \textit{https://archive.org/details/stackexchange} in XML format.
We started with 330,705 candidate visual questions. After removing visual questions without an accepted answer, it resulted in 165,766 visual questions. As mentioned in the main paper, we then conducted two filtering steps. After removing visual questions with scores of 0 or less for either the question or answer, we had 119,177 visual questions. Next, after removing visual questions with multiple images, we had 85,573 visual questions.  Subsequently, removing visual questions with visual answers, left 65,849 visual questions.   Examples of filtered visual questions with visual answers and multiple images are shown in Figures \ref{fig:visualanswer} (a), and (b), respectively. Finally, after removing examples for which the image could not be downloaded from the provided link, we got to our final dataset of 64,696 visual questions. For data format consistency, we converted all images to png format. 
 
\subsection{User Intention Taxonomy and Definitions}
\textbf{User Intention Taxonomy.} We first brainstormed 11 free-form user intention categories.  To do so, we solicited help from GPT-3.5 by having it indicate via a zero-shot prompt the user intentions when given a question and context.  We identified an initial 11 free-form user intention categories, which are shown in the second column of Table \ref{Table:taxonomy}. We then refined these categories based on the frequency of their occurrence in the literature on user intention (shown in the third to the last column in Table \ref{Table:taxonomy}) for related fields such as question answering and search queries~\cite{bolotova2022non,cambazoglu2021intent,ignatova2009annotating,harper2010question,toba2014discovering,chen2012understanding,kofler2016user,liu2015understanding} to the following 7 categories: advice, evidence, identification, instruction, opinion, reason,  and verification. We then drafted the definitions for each taxonomy, mainly by adapting definitions from \cite{cambazoglu2021intent,bolotova2022non, harper2010question}. 

We then finalized the taxonomy with definitions via four rounds of annotator analysis on a total of 105 visual questions.  Specifically, two annotators (authors) tagged the category for each visual question. Any disagreements that arose were discussed and resolved, and the definitions associated with each taxonomy category were adjusted accordingly. We observed slight agreement accuracy \footnote{The agreement accuracy is calculated by dividing the number of agreements between two annotators by the total number of examples annotated.} improvement and Cohen's kappa agreement improvement after each round. Specifically, from round 1 to round 4, agreement accuracy improved from 53.3\% to 87.5\% and Cohen's kappa agreement improved from 45.24\% to 84.69\%. This process culminated in our final taxonomy and their definition, which includes the following eight categories mentioned in the main paper: \emph{advice, evidence, identification, instruction, opinion, reason, verification}, and \emph{other}.  

As shown in Table \ref{Table:taxonomy}, our final taxonomy is different from ~\cite{bolotova2022non,cambazoglu2021intent,ignatova2009annotating,harper2010question,toba2014discovering,chen2012understanding,kofler2016user,liu2015understanding} as they focus solely on text-based question or search queries whereas our work centers on visual questions. \cite{brady2013visual} also explores user intentions for authentic visual questions for blind individuals. 
However, blind people's user intents of visual questions significantly differ from that of online users' visual questions. In fact, only one of the user intents identified for blind individuals (i.e., identification) overlaps with the user intents we have identified for our dataset. 

\paragraph{\bf{Final user intention taxonomy and definitions.}}
We show the final user intention taxonomy and definitions as follows: 
\begin{itemize}[itemsep=0.1em,leftmargin=*]
    \item \textbf{Verification}: These fact-checking questions aim to confirm or refute a hypothesis through affirmative answers (yes/no or multiple choice options). Since this question is based on fact, it cannot include subjective questions which should be classified as Opinion. 
   \item \textbf{Identification}: The expected answer for these questions is a named entity or object identification. The answer is objective.
   \item \textbf{Reason}: These questions require answers that explain causes underlying a particular action or event. Most "why" questions fall into this category. 
   \item \textbf{Evidence-based}: This category primarily includes questions that ask for the definition, features, description, or process of a concept, idea, object, event, or one of its attributes. It also covers translation questions of seeking the meaning of a sentence or word. If a question can be classified as Verification, it should not be classified as Evidence-based. 
   \item \textbf{Instruction}: These questions typically involve "how to do" inquiries, seeking instructions, guidelines, or procedures for a specific action in real life. 
   \item \textbf{Advice}: This category includes questions where the user seeks personalized guidance on a specific topic. Advice questions differ from Instruction questions in that they expect subjective recommendations, while Instruction questions seek objective, step-by-step processes for performing a specific action. Advice questions may also involve finding resources or seeking better translations of sentences/words. Additionally, this category can include questions where users are looking for ideas or comments on how to improve an existing solution. If a question can be classified as Instruction, it should not be categorized as Advice.
    \item \textbf{Opinion}: These questions aim to elicit subjective opinions on a topic of interest (e.g., "what do you think about" or "is X good/bad"). It might include religious questions. This category excludes Advice questions, where the focus is on the user asking the question.
    \item \textbf{Other}: Other.
\end{itemize}

\subsection{User Intention Annotations }
\paragraph{\bf{Hiring Crowdworkers.}}
We hired three crowdworkers from Amazon Mechanical Turk to perform our task. For quality control, we only accepted workers located in the United States who had completed more than 500 Human Intelligence Tasks (HITs) with over a 95\% approval rating.   
\begin{table*}[hb!]
\centering
\caption{Eight user intent taxonomy categories: one from our dataset, one for free-form intent (derived through GPT-3.5 bootstrapping and manual labeling), and six from existing research.  Our taxonomy specifically addresses user intentions for visual question answering, which is in contrast to the taxonomy from \cite{cambazoglu2021intent} which focuses on Bing query search and others ~\cite{harper2010question,bolotova2022non,toba2014discovering,ignatova2009annotating,fu2016music} that are dedicated to community question answering.  }
\scalebox{0.59}{
\begin{tabular}{@{}llllllll@{}}
\toprule
Ours          & free-form intent & Ignatova (2009) \cite{ignatova2009annotating} & Harper (2010)\cite{harper2010question} & Toba (2014) \cite{toba2014discovering} & Cambazoglu(2021) \cite{cambazoglu2021intent} & Bolotova (2022) \cite{bolotova2022non} & Fu (2016) \cite{fu2016music}\\ \midrule
Instruction    & How  & Procedural                                              & 
Prescriptive                                       & Proceure                                       & Process                                                 & Instruction                                 & 
Instruction    \\
Evidence-based         & Comprehend fact&General Info need                                & Factual                                           & Factoid                                          & Description                                              & Evidence-based               & Factual                    \\
Reason         & Why &Causal                                                  & -                                                 & Reason                                         & Reason                                                  & Reason                               &-      \\
Verification   & Validate &Verification                                             & -                                                 & Yes/No                                          & 
Verification                                            & Debate                        &-       \\
Opinion       & - &Disjunctive    & Disapproval/Quality                                       & Opinion                                        & Opinion                                                 & -                                   &Opinion  \\
Identification & Recognize&Concept completion                                      & Identification                                    & -                                               & Entity                                                  & -                                          &Identifying resources \\
Advice         & Advise & -                                                       & Advice                                            & -                                               & Advice                                                  & Experience            &Recommend/Solution         \\

-              & Prove &Quantification                                          & -                                                 & -                                               & Quantity                                                & -                                                \\
-          & Compare &Comparison                                              & -                                          & -                                               & -                                                       & Comparison                                       \\
-      & - & Definition                                              & -                                                 & Definition                                      & -                                                       & -                                                \\

-   & Find& -                                                       & -                                                 & -                                               & Resource                                               & -                                                \\

-              & Language& -                                                       & -                                                 & -                                               & Language                                                & -                                                \\
-              &-& -                                                       & -                                                 & -                                               & Temporal                                                & -                                                \\
-              & -&-                                                       & -                                                 & -                                               & Calculation                                             & -                                                \\
-              & &-                                                       & -                                                 & -                                               & Attribute                                               & -                                                \\
-              &- &-                                                       & -                                                 & -                                               & List                                                    & -                                                \\
-              & -&-                                                       & -                                                 & -                                               & Weather                                                & -                                                \\
-              & &-                                                       & -                                                 &f -                                               & Location                                                & -                                                \\              & Explain &-                                                       & -                                                 & -                                               & -                                              & -                                                \\-      & - &-                                                       & -                                                 & -                                               & -                                              & -                                    & Request research \\   \\Other      & - &-                                                       & -                                                 & -                                               & -                                              & -                                                \\
\bottomrule
\end{tabular}}

\paragraph{\bf{User Intention Annotations Task Design.}}
Crowdworkers labeled one primary intent per visual question when shown the question, image, context, and answer.  For the final label, we used the majority vote label per VQA from 3 crowdsourced labels per visual question.

\label{Table:taxonomy}
\end{table*}
\paragraph{\bf{Annotation Task Design and Collection. }}
We provided instructions, taxonomy definitions, and two examples per user intention category. The annotation task interface showed for each visual question the question, context, and answer. 

To facilitate collecting high-quality results, we provided each crowdworker with one-on-one Zoom training. We also provided a qualifying annotation test that all workers passed to verify they understood the instructions. 

To enable the assessment of high-quality results, we collected two user intention annotations from two independent workers (worker A and B) per VQA instance (i.e. image-question-context-answer) for 105 randomly selected VQAs, one from each of 105 topics. We then detected if their annotations matched. We found that 25 out of 105 didn't match. Thus, we hired a third worker (worker C) to break the tie by instructing that individual to select one of the two provided user intentions. We paid \$40 Amazon Gift Card to workers A and B and \$5 Amazon Gift Card to worker C.    
 
\paragraph{\bf{Instructions for User Intention Annotation.}}
The following are the instructions provided for workers A and B: 
\begin{itemize}[itemsep=0.1em,leftmargin=*]
    \item Read definitions and two examples for each category.
    \item  We will present to you 105 questions. Each question contains a question, context, and a reference answer. Please read them and categorize the primary user intention according to step 1.   
    \item If there are any non-English questions, or if the questions contain many proper nouns that are difficult to interpret, you can use a translation tool to translate them into your native language.
    \item Fill in the user intent category for each  question in the provided spreadsheet.
\end{itemize}

\subsection{Potential Benefits of User Intention in VQA}
Though we only provide an initial exploration of user intention in the main paper, we discuss the potential usage of user intention here to encourage future extensions. Overall, understanding and identifying user intent can potentially help create human-centered VQA models for better user experience.  First, it facilitates the creation of datasets with better answers tailored to user needs. Second, analysis of intention prevalence in real-world scenarios—examining alongside model performance for each type can help developers in prioritizing their efforts. Third, a model with the ability to understanding user intent also can potentially provide answers more directly meeting users needs, rather than related, true responses not meeting users' needs. For instance, when asked a visual question such as "why doesn't the bulb work," users may prefer practical solutions to fix the bulb rather than just reasons and explanations. Without recognizing user intentions, models might only offer reasons. 

\subsection{Potential Limitations and Societal Impact of VQAonline Dataset.}
Despite the VQAonline dataset's inclusion of multilingual VQAs and coverage of users' generated visual questions from various countries, we highlight that this dataset may poorly represent people in poverty-stricken areas since VQAs are posted by Stack Exchange users with access to mobile devices and Internet. Additionally, the geographic representation of Stack Exchange is unbalanced, with more from countries like US, India, Europe and less from countries like Egypt. 

\section{Dataset Analysis}
We supplement the main paper by comparing the sources of the visual questions for eight datasets in Table \ref{tab:SourceDatasetCompare}. Only our dataset and VizWiz-VQA dataset are sourced from authentic use cases with both the images and visual questions coming from real users. 











\begin{table*}[!b]
\centering

\caption{Details about the six benchmarked models' model configuration and training data. CC$^{*}$ comprises datasets from COCO~\cite{chen2015microsoft}, CC3M~\cite{sharma2018conceptual}, and CC12M~\cite{changpinyo2021conceptual}. CC stands for Conceptual Caption~\cite{sharma2018conceptual,changpinyo2021conceptual}; VG stands for Visual Genome~\cite{krishna2017visual}; CY stands for COYO-700M~\cite{byeon2022coyo}; L400 stands for LAION 400M~\cite{schuhmann2021laion}; SBU~\cite{ordonez2011im2text} contains 1 million images with captions. LLaVA-I stands for 158K multimodal instruction-following data in LLaVA~\cite{liu2023visual}. QA$^{*}$ stands for 13 question-answering datasets in InstructBLIP~\cite{instructblip}.}
\scalebox{0.852}{
\begin{tabular}{l|lll|ll|ll}
\toprule

                &  \multicolumn{3}{c} {Model Configuration} & \multicolumn{2}{|c} {Img-text Data} & \multicolumn{2}{|c} {Visual Instruction Data}\\ 
\cline{2-8} 
Models                & Img Encoder & Lang Encoder  & Adapter &Source &Size&Source&Size\\ 
\midrule
BLIP2 ~\cite{li2023blip}  &  ViT-g/14             &      FlanT5XL         &   Q-Former &CC$^{*}$, VG, SBU, \textcolor{blue}{L400}& 129M& -&-\\
 MiniGPT-4~\cite{zhu2023minigpt}             &    BLIP2-VE   &    Vicuna 13B& FC layer &CC, SBU, \textcolor{blue}{L400}& 5M &CC+ChatGPT&3.5K \\ 
LLaVA~\cite{liu2023visual}        &   ViT-L/14      &   LLaMA 13B     & FC layer &CC3M & 595K&LLaVA-I&158K\\ 
mPLUG-Owl~\cite{ye2023mplugowl}           & ViT-L/14          &     LLaMA 7B &   LoRA &CC$^{*}$, CY, \textcolor{blue}{L400}&204M&LLaVA-I&158K\\ 

InstructBLIP~\cite{instructblip} &        ViT-g/14      &FlanT5XL   & Q-Former    &-&-&QA$^{*}$& 16M\\ 
GPT-4V ~\cite{OpenAI2023GPT4TR}        & Unknown         &   Unknown     & -  & Unknown &-&-&- \\

\bottomrule
\end{tabular}}
\label{table:bm}
\end{table*}
\begin{table*}[!b]
\centering
\caption{Details about the six benchmarked models' language encoders and image encoders. L400, L2B, and C4 are all derived from Common Crawl. (L400=LAION-400; L2B=LAION-2B; C4=Colossal Clean Crawled Corpus).}
\scalebox{0.852}{
\begin{tabular}{lp{12cm}}
\toprule

Encoder            & Source\\ 
\midrule
ViT-L-14~\cite{radford2021learning}     & \textcolor{blue}{L400~\cite{schuhmann2021laion}}  \\ 
ViT-g/14~\cite{fang2023eva}      &   \textcolor{blue}{L2B~\cite{schuhmann2022laion}}    \\

\hdashline
FlanT5XL~\cite{chung2022scaling}& Finetuned based on T5~\cite{raffel2020exploring} while T5 was trained with \textcolor{blue}{C4}~\cite{raffel2020exploring}. Finetuned with 69 Multi-task finetuning with instructions (Muffin) datasets~\cite{wei2021finetuned, byrne2019taskmaster,anantha2020open,dai2022dialog,yasunaga2020graph,li2022competition}, 55 datasets from T0-SF~\cite{sanh2021multitask}, and 375 natural instructions v2 (NIV2) datasets~\cite{wang2022benchmarking} and 9 Chain-of-Thought (CoT) datasets derived from ~\cite{cobbe2021training,geva2021did,camburu2019make,ling2017program,onoe2021creak,aggarwal2021explanations,khot2020qasc,lamm2021qed,wang2019does}\\
LLaMA~\cite{touvron2023llama1}     &\textcolor{blue}{CommonCrawl}, \textcolor{blue}{C4}~\cite{raffel2020exploring}, Github, Wikipedia, Gutenberg and Books3~\cite{gao2020pile}, ArXiv~\cite{lewkowycz2022solving}, and QAs from \textcolor{red}{StackExchange's 28 largest websites}. \\
Vicuna     & Finetuned based on LLaMA~\cite{touvron2023llama1} with user-shared conversations, so also with \textcolor{blue}{CommonCrawl}, \textcolor{blue}{C4}~\cite{raffel2020exploring}, Github, Wikipedia, Gutenberg and Books3~\cite{gao2020pile}, ArXiv~\cite{lewkowycz2022solving}, and QAs from \textcolor{red}{StackExchange's 28 largest websites}.\\

\bottomrule
\end{tabular}}
\label{table:encoder}
\end{table*}

\section{Algorithm Benchmarking}

\paragraph{\bf{Architectures.}}
Details about each of the six benchmarked models are provided in Table \ref{table:bm} and Table~\ref{table:encoder}.  Specifically, we report each model's image encoder, language encoder,  adapter, and their training data \cite{xu2023lvlm,OpenAI2023GPT4TR,zhu2023minigpt,instructblip,liu2023visual,ye2023mplugowl,li2023blip}.  Dataset sources that have data contamination are marked in red and sources suspected with data contamination are marked in blue~\footnote{All models have data coming from LAION400 (L400), LAION2B(L2B), or C4, which are in turn derived from Common Crawl that is known to contain random web pages (and so could have come from Stack Exchange). GPT-4V training details are not publicly-available but are believed to involve internet data.}.


\paragraph{\bf{Model Implementations.}}
For GPT-4V, we used the gptv-2023-07-01-preview version. 

For one-shot settings, we created the prompts by adjusting prompts from mPLUG's official repository and Azure's few-shot prompting examples for groundedness evaluation. The prompts we used is exemplified in Figure~\ref{fig:oneshotprompt}.
\begin{promptbox}
"The following is a conversation between a curious human and AI assistant. The assistant gives helpful, detailed, and polite answers to the user's questions.\\

\#\# Example Task\\
Human: <image\_example>\\
Human: \{question\_example\} + \{context\_example\}\\
AI: \{reference\_answer\}"\\

\#\# Actual Task:\\
Human: <image>\\
Human:  \{question\} + \{context\}\\
AI: "
\end{promptbox}
\captionof{figure}{The prompt for one-shot setting. }
\label{fig:oneshotprompt}

As mentioned in the main paper, to complement five popular evaluation metrics, we also introduce a new evaluation metric based on LLaMA2~\cite{touvron2023llama}. We tested four different prompts for it to assess the correctness of model-generated answers: prompting to output continuous scores from 0 to 1, discrete scores (0, 0.5, 1), following~\cite{zheng2024judging}, or Azure groundness evaluation examples.  We selected the best one from our preliminary analysis, which is shown in Figure~\ref{fig:llama2prompt}.
    
\begin{promptbox}
System: "You are a helpful AI assistant. You will be presented with a REFERENCE ANSWER and a PREDICTED ANSWER. Your task is to rate the correctness of the PREDICTED ANSWER. 
Chose one of the following rating: 0 (Totally Wrong), 0.5 (Partially Correct), or 1 (Totally Correct).

Just complete the last space of the correctness score."\\

User: "REFERENCE ANSWER: \{Reference\} \\
PREDICTED ANSWER: \{Prediction\} \\
Score: "
\end{promptbox}
\captionof{figure}{The prompt for LLaMA2 metric.}
\label{fig:llama2prompt}

\paragraph{\bf{Subset Creation. }}
As mentioned in the main paper, due to GPT-4V model requiring a fee for their usage~\cite{gpt4vpricing2023}, and due to the computational and financial cost of running LLaMA2~\footnote{As mentioned in the main paper, LLaMA2 is impractical at scale due to computing time. For example, it takes 4 days for 4 Quadro RTX 8000 GPUs to compute LLama2 score for one model on  the entire VQAonline dataset. Considering we have 13 models to evaluate (six baseline models, mPLUG-Owl models with three different input types, and mPLUG-Owl models with four different one shot-settings), such evaluation would take 52 days.}, we only evaluate GPT-4V and LLaMA2 on a subset of data.

We created this subset by randomly selecting 20 VQAs from each of the 105 topics in the test set, except for the 17 topics containing less than 20 examples where we used all available VQAs.  The resulting subset contains 1903 VQAs. Another reason for creating the subset with random sampling is due to the original dataset displaying a long-tail distribution across different topics. 

\subsection{\bf{Results on Subset. }}
\paragraph{\bf{Overall Results on Subset. }}
Results are shown in Table~\ref{table:bmresultsubsetall}. As mentioned in the main paper, GPT-4V achieves the best performance on the subset, based on five of the six evaluation metrics. Other models have the same rankings as that shown in the main paper. 

\paragraph{\bf{mPLUG-Owl Model in One-shot Setting, Evaluated with LLaMA2 Metric. }} 
We explore the mPLUG-Owl model in one-shot-setting on subset, evaluated with LLaMA2 metric. Results are shown in Table~\ref{table:bmresultSubsetOneshot}. It strengthens our finding in the main paper that none of the one-shot examples enhance performance for mPLUG-Owl and that exemplars with matching topic tags consistently outperform randomly chosen ones.

\begin{table}[t!]
\centering
\caption{Performance of VLMs on the \textbf{subset} with 1,903 random samples from VQAonline dataset with respect to six (ROUGE-L, METEOR, BERTscore, CLIP-S, RefCLIP-S, and LLaMA2) for the zero-shot setting. As shown, GPT-4V is the best-performing model in the zero-shot setting.    } 
\begin{tabular}{p{3cm}llllll}
\toprule
Models                &ROUGE-L &  METEOR  & BERTscore & CLIP-S & RefCLIP-S & LLaMA2 \\ 
\midrule
 GPT-4V\cite{OpenAI2023GPT4TR}&\textbf{0.16 }& \textbf{0.12} &\textbf{0.76}& 0.70 & \textbf{0.75} & \textbf{0.70} \\
mPLUG-Owl \cite{ye2023mplugowl}    &  0.14    &     0.10   & 0.75      & 0.71 & 0.74& 0.62\\ 
LLaVA   \cite{liu2023visual}  &       0.14  &    0.08  &    0.75     &\textbf{0.72} & 0.74& 0.62 \\ 
  MiniGPT-4 \cite{zhu2023minigpt} &    0.13     &    0.08  &    0.74     &0.69& 0.73& 0.59\\ 
InstructBLIP  \cite{instructblip} &   0.09      &   0.06   &    0.69     &0.68&0.71 & 0.59\\
 BLIP2 \cite{li2023blip}    &  0.07       &     0.04 &        0.69 &0.65&0.70 &  0.53\\
\bottomrule
\end{tabular}
\label{table:bmresultsubsetall}
\end{table}

\begin{table}[]
\caption{Performance of mPLUG-Owl on the \textbf{subset} of VQAonline dataset with respect to the LLaMA2 metric with four different one shot settings. }
\centering
\begin{tabular}{p{5cm}l}
\toprule
Models                &LLaMA2 \\ 
\midrule
mPLUG-1shot-Random-noImg                & 0.609 \\ 
mPLUG-1shot-MatchedTopic-noImg             & 0.617 \\ 
mPLUG-1shot-Random      &  0.608\\ 
mPLUG-1shot-MatchedTopic       & 0.617 \\ 
\bottomrule
\end{tabular}
\label{table:bmresultSubsetOneshot}
\end{table}
\begin{table}[t!]
\caption{Fine-grained analysis of the top-performing VQA models on the \textbf{subset} of the VQAonline dataset, GPT-4V and mPLUG-Owl,  when fed different input types (Q+C+I, C+I, Q+C, Q+I), evaluated with LLaMA2 metric.} 
\centering 
\begin{tabular}{p{5cm}l}
\toprule
Models                &LLaMA2  \\ 
\midrule
mPLUG (Q+C+I)    & \textbf{0.62}\\
mPLUG (C+I)     & 0.61\\
mPLUG (Q+C)    & 0.60\\
mPLUG (Q+I)  & 0.58\\
\hdashline
GPT-4V (Q+C+I)& \textbf{0.70}\\
GPT-4V (C+I) & 0.69\\
GPT-4V (Q+C) & 0.69\\
GPT-4V (Q+I) & 0.62\\
\bottomrule
\end{tabular}
\label{table:fine-grained-analysis-subset}
\end{table}

\begin{table}[]
\caption{Fine-grained analysis of the top-performing VQA models on the \textbf{subset} of the VQAonline dataset, GPT-4V and mPLUG-Owl,  with respect to each of the five VQA super-category types, evaluated with LLaMA2 metric.} 
\centering 
\begin{tabular}{p{5cm}l}
\toprule
Models                &LLaMA2 \\ 
\midrule
mPLUG (Science)    & 0.69\\
mPLUG (Life\&Arts)   & 0.59\\
mPLUG (Culture\&Recreation)  &0.60\\
mPLUG (Business)   & \textbf{0.72}\\ 
mPLUG (Professional)   & 0.66\\
\hdashline
GPT-4V (Science)    & 0.75\\
GPT-4V (Life\&Arts)   & 0.67\\
GPT-4V (Culture\&Recreation)  & 0.67\\
GPT-4V (Business)   & \textbf{0.76}\\ 
GPT-4V (Professional)   &  0.74\\
\bottomrule
\end{tabular}
\label{table:supertopics-subset}
\end{table}

\paragraph{\bf{Anaysis With Respect to Input Types, Evaluated with LLaMA2 Metric. }} 
We then analyze the predictive power of each input of the top-performing models, mPLUG-Owl and GPT-4V on subset evaluated with LLaMA2 metric. Results are shown in Table~\ref{table:fine-grained-analysis-subset}. The findings from LLaMA2 metric strengthens the findings we discussed in the main paper that (1) best-performing model is with all information (Q+C+I), (2) the worst-performing model is the model lacking context (i.e., Q+I) for both GPT-4V and mPLUG-Owl models, and (3) the context is the most valuable information source for arriving at the target answer, while the predictive powers of questions and images are nearly negligible.

\begin{figure}[t!]
    \centering
    \includegraphics[width=0.6\textwidth]{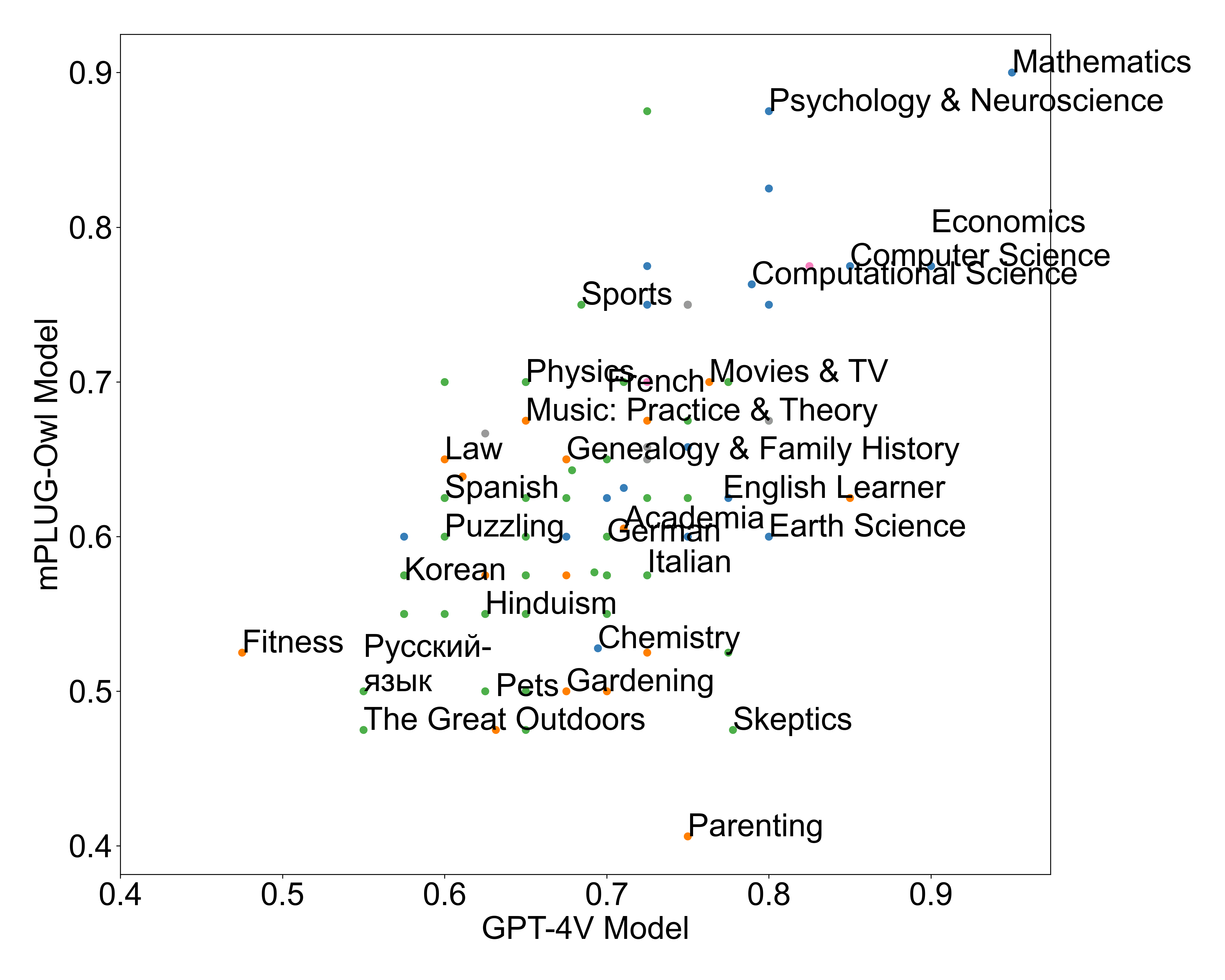}
  \caption{Performance of mPLUG-Owl and GPT-4V on the subset of VQAonline, for each of 105 topics with their five super-categories represented in 5 different colors.  Results are shown with respect to LLaMA2 metric.}
    \label{fig:metricsCorrelationgptv}
\end{figure}

\paragraph{\bf{Analysis With Respect to VQA Topics. }} 
We next analyze the influence of VQA topic on the top-performing models, mPLUG-Owl and GPT-4V on the subset, with respect to the five super-categories in Table~\ref{table:supertopics-subset} and with respect to 105 topics, evaluated with LLaMA2 metric in Figure~\ref{fig:metricsCorrelationgptv}. 

Overall, the models perform best in the Business category and worst in the Life \& Arts category. Note that this does not contradict our findings in the main paper, as models still perform well in Science and poorly in Culture \& Recreation. We suspect that the differences between this and the main paper's findings are due to the distribution differences (different proportion of each topic in each super-topics) of the subset compared to the entire dataset. To support our hypothesize, we also calculated METEOR, and BERTscore, which are highly correlate with human judgments and observed the same trend as LLaMA2 metric on subset. 

\begin{figure}[t!]
    \centering
    \includegraphics[width=1\textwidth]{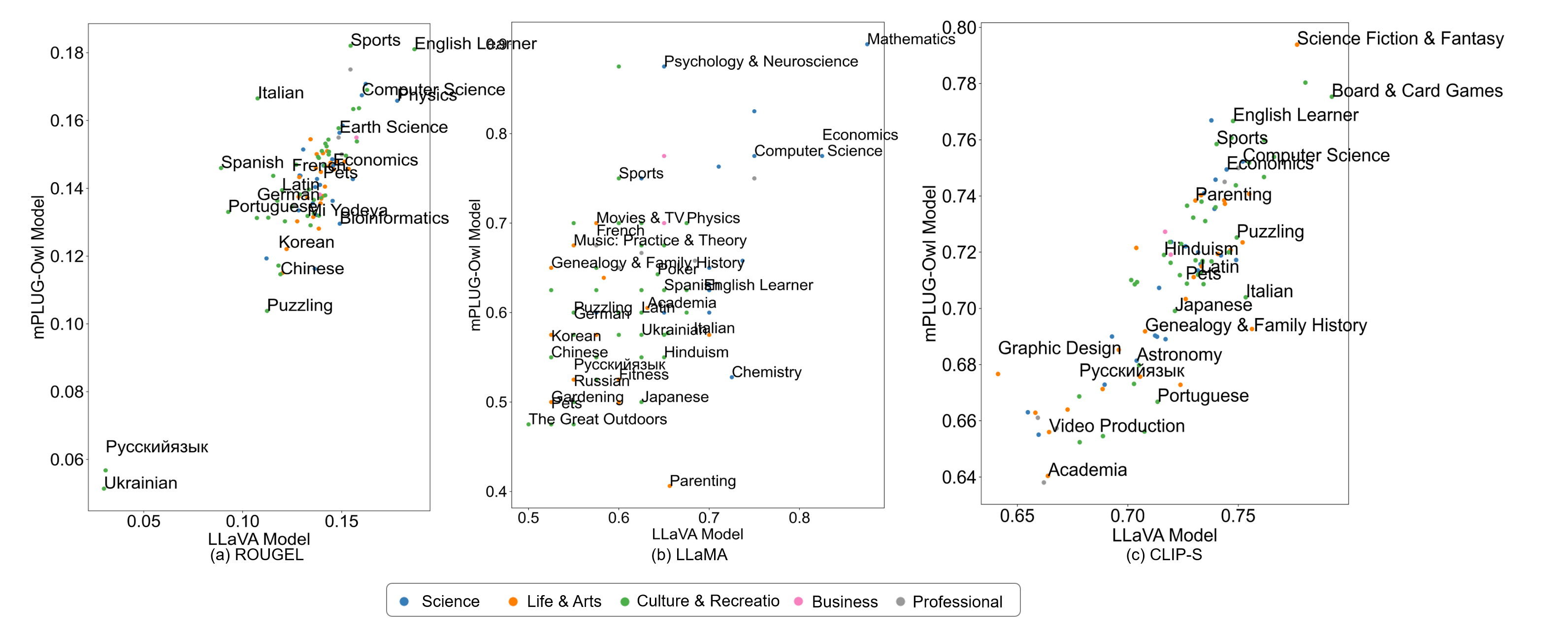}
  \caption{Performance of mPLUG-Owl and LLaVA on the VQAonline, for each of 105 topics with their five super-categories represented in 5 different colors.  Results are shown with respect to three evaluation metrics: (a) ROUGEL, (b) LLaMA2, and (c) CLIP-S. Note that we only report LLaMA2 metric on the subset. For visualization simplicity, we show text labels only for topics with identified interesting trends. We omitted "language" for each language topic and omitted topics with less than 10 data points. We also shortened some topics' names, such as "Gardening" instead of "Gardening \& Landscaping", and "Fitness" instead of "Physical Fitness".}
    \label{fig:metricsCorrelationAll}
\end{figure} 

To complement the main paper, we also reported mPLUG-Owl and LLaVA across 105 topics with three other metrics in Figure~\ref{fig:metricsCorrelationAll}. Overall, it strengthens our findings in the main paper regarding what topics are relatively easy (\eg Mathematics, Economics, and Computer Science) and hard (\eg Pyccknn, Hinduism, Puzzling). Besides, LLaMA2 indicates that some Culture \& Recreation topics (e.g.,  "The Great Outdoors") and some Life \& Arts topics (e.g., "Pets", "Gardening \& Landscaping", "Physical Fitness") are relatively hard for GPT-4V, mPLUG-Owl, and LLaVA models.

\paragraph{\bf{Analysis With Respect to Image Necessity.}}
We conducted a small scale model performance analysis for the Image Necessity subset, created in section ``VQAonline Importance of Images". The top-performing model's results are shown in Table~\ref{table:Importance}.


\paragraph{\bf{Analysis With Respect to Text Presence.}}
We conducted text spotting model to determine the presence of text within the images. We utilized the state-of-the-art text spotter model DeepSolo to detect text for all images in our dataset. We found 76\% (i.e., 48,161 images) contain text. The metadata is shared at: https://vqaonline.github.io/.  The top-performing model, mPLUG-Owl, performs better overall for images with text, as shown in Table~\ref{table:Importance}.  While interesting, we want to highlight that the presence of text does not necessarily mean that questions ask about that text.

\begin{table}[b!]
\centering
\small
\begin{tabular}{lccccc}
\toprule
Data & ROUGE-L & METEOR & BERTscore & CLIP-S & RefCLIP-S\\
\midrule 
Necessary & 0.146 & 0.096 & 0.748 & 0.713 &0.738 \\
Not Necessary&0.139 & 0.102  &0.753  &  0.692 &0.734 \\
\hdashline
With Text  & 0.144 & 0.12  & 0.767  & 0.725 &   0.757\\
Without Text&  0.143&  0.111& 0.753  & 0.723&0.740  \\
\bottomrule
\end{tabular}
\caption{Fine-grained analysis of the top-performing model, mPLUG-Owl with respect to five evaluation metrics when analyzing with respect to whether the image is necessary to answer the question and whether text is in the image. } 
\label{table:Importance}
\end{table}

\subsection{Experiments with Domain Experts}

\paragraph{\bf{Topic Selection.}}
We chose the 10 topics based on initial manual selection and expert availability. Initially, we manually identified 20 candidate topics for human evaluation. This selection considered diversity including using three of the seven topics from~\cite{xu2023critical} as well as topics on math (i.e., stats), language (i.e., Chinese), and daily life (e.g., Gardening and Fitness). The topics' frequency span from very frequent (Stats, Gardening, and Music) to moderate frequency (Economics, AI) to less frequent (Law). Subsequently, we selected ten topics based on the availability of domain experts to represent each area.  We acknowledge that these 10 topics might not fully present the entire dataset, as it is a chicken-and-egg problem: we don't know the difficulty level of the 105 topics beforehand and thus have to first conduct a human evaluation on a few topics to decide which metrics are most human-aligned for quantitatively evaluating the entire dataset.  

\paragraph{\bf{Domain Expert Hiring.}}
We hired ten domain experts to represent each of the ten fields (topics). To guarantee English proficiency, as most of our visual questions are written in English, we only accepted experts located in the United States.

\begin{figure*}
         \centering
     \includegraphics[width=1\textwidth]{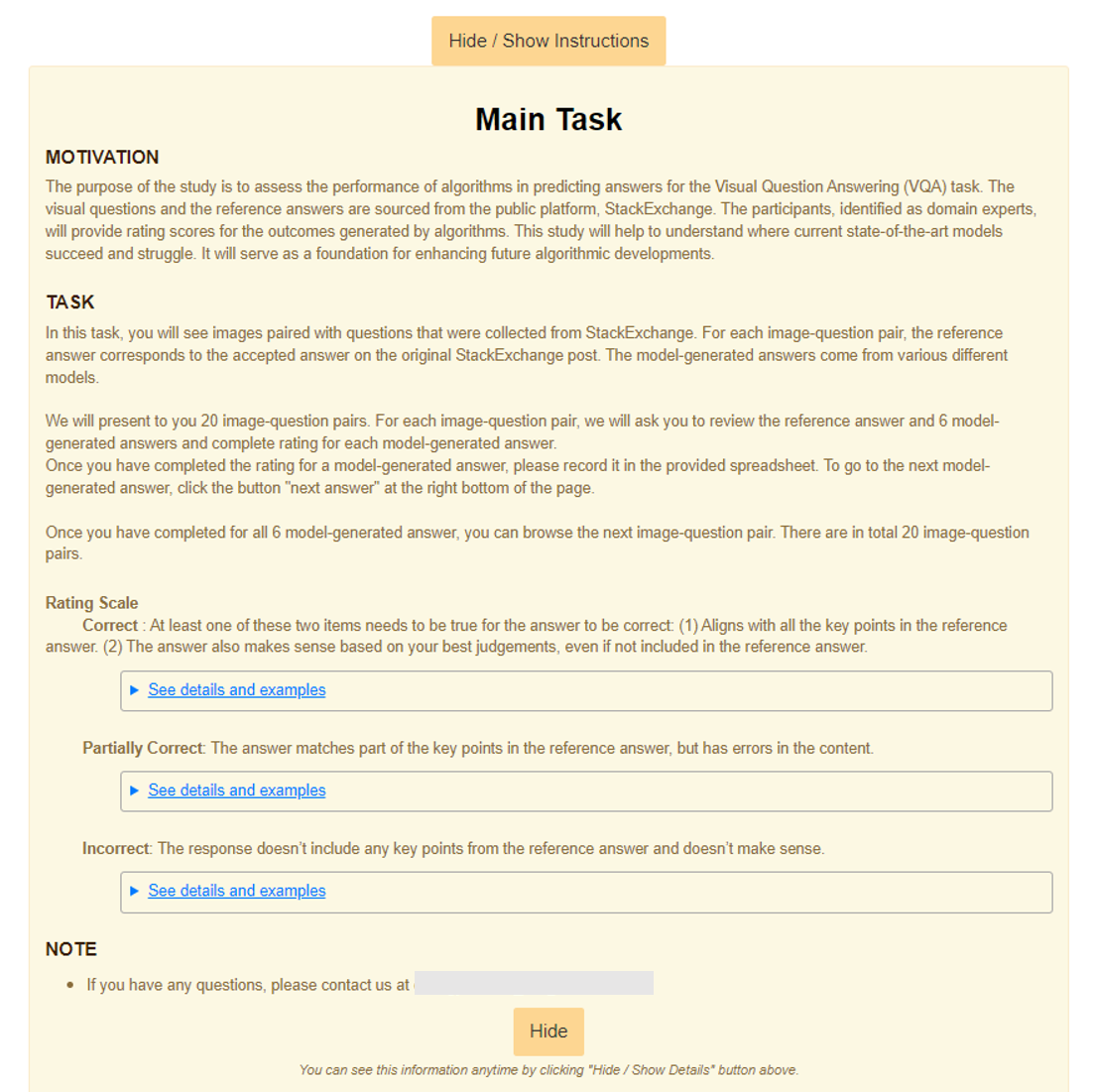}
        \caption{Instructions for our domain expert annotation task. }
        \label{fig:instruction}
\end{figure*}

\begin{figure*}
     \centering
     \includegraphics[width=1\textwidth]{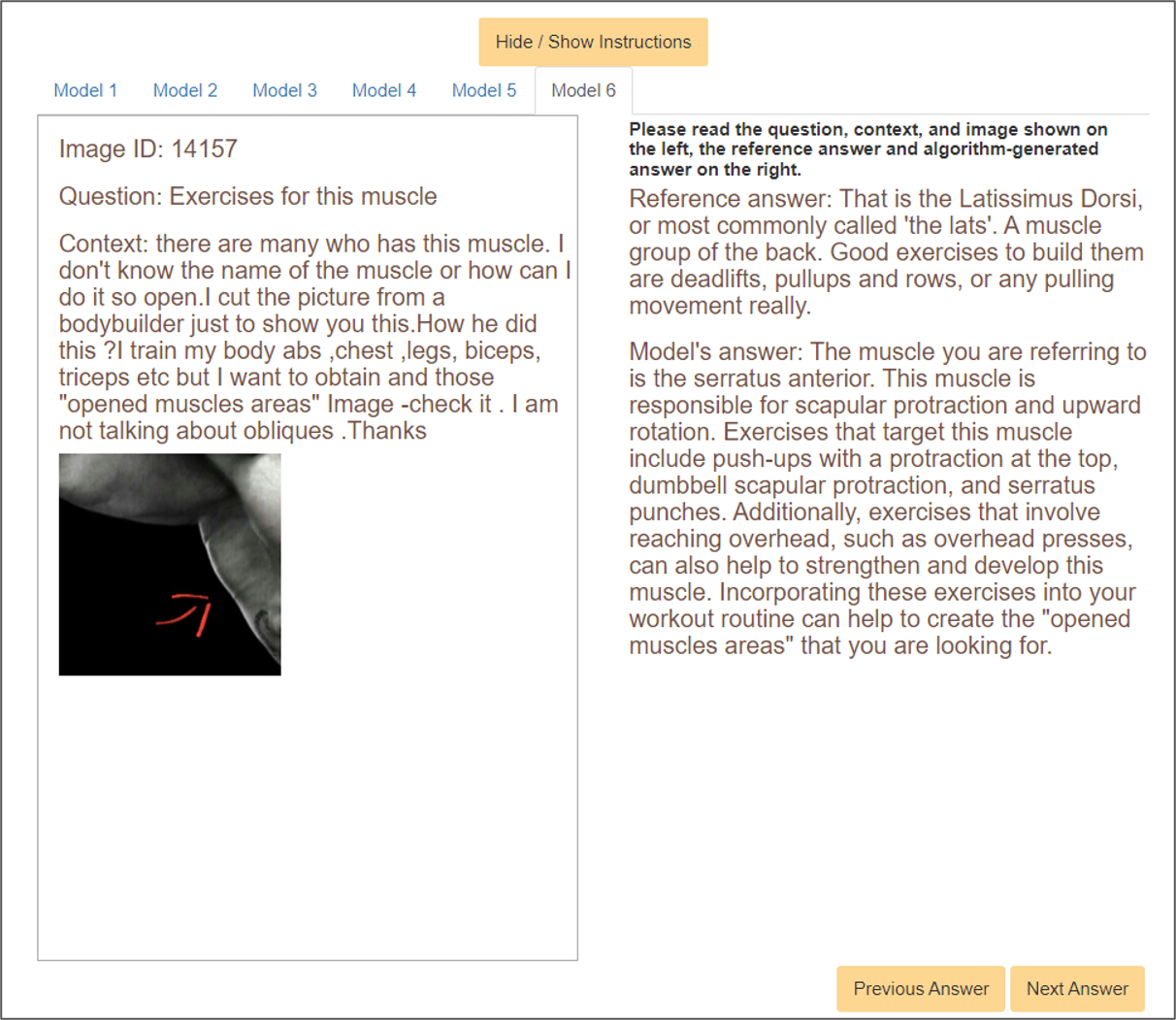}
         \caption{Screen shot of the annotation task interface. Experts recorded their results in spreadsheets. }
         \label{fig:userinterface}
\end{figure*}

\paragraph{\bf{Annotation Task Interface.}}
We show a screenshot of the annotation instructions in Figure \ref{fig:instruction} and task interface in Figure \ref{fig:userinterface}. The link to the code for our web-based annotation tool is available at https://github.com/VQAonline/VQAonlineVisualization.

\paragraph{\bf{Data Collection Quality Control Mechanism.}}
We provided a one-on-one training session with each domain expert to introduce the task and instructions. Afterwards, we gave each expert our contact information so that they could contact us with any questions about their tasks and receive feedback quickly. We also inspected the time each expert spent on each visual question as a proxy for assessing whether the hired domain experts were finishing the task by quickly selecting random options. The median and mean times are 1.4 minutes and 1.29 minutes respectively for assessing each model answer per visual question. We compensated each domain expert with a \$75 Amazon Gift Card, resulting in an average hourly wage of \$26/hour. 

\paragraph{\bf{Spearman Correlation. }}
To supplement the main paper, we also report the Spearman correlation scores for each evaluation metric:  GPT-4 (1 - correlation, 0.000 - statistical significance), ROUGEL (1, 0.000), METEOR (0.943,0.005), BERTScore (0.829, 0.042), RefCLIP-S (0.771, 0.072), and CLIP-S (0.600,0.208). Spearman correlation rankings for the models follow the order from human judgments, reinforcing the main paper's findings that reference-based metrics and human judgments are highly correlated.

\paragraph{\bf{Correct, Partially Correct, and Incorrect Examples From Expert Evaluation. }}
We show examples of expert annotations for the six benchmarked models in Tables \ref{tab:ve00}, \ref{tab:ve11}, \ref{tab:ve12}, \ref{tab:ve31}, \ref{tab:ve32}, \ref{tab:ve41}, spanning those that are correct (\ref{tab:ve00}), partially correct (Table \ref{tab:ve11} and Table \ref{tab:ve12}), and incorrect (\ref{tab:ve31}, \ref{tab:ve32}, \ref{tab:ve41}). 

One mechanics's example shows where most of the models can answer correctly for a closed-ended visual question related to recognition (Table \ref{tab:ve00}). 

Two economics examples show where the models partially fail for analyzing an \emph{infographic} and requiring \emph{specific domain knowledge} (Table \ref{tab:ve11} and Table \ref{tab:ve12}). These highlight that models labeled "partially correct" can occur because of correctly answered closed-ended questions with incorrect explanations or insufficient explanations matching the reference answer. For example, GPT-4V, mPLUG-OWl, MiniGPT-4, and LLaVA answer "yes" but don't match any key points of the reason in the reference answer and have factual errors in their given reasons \footnote{Of note, the errors in all models' answers to the economic VQ examples are highlighted in red by a Ph.D. student in economics. }. InstructBLIP answered "yes" without providing any reason, but simply rephrased the context. For each example, key points in the reference answer were either unsatisfactorily conveyed in the model's answer or were altogether absent. 

We also show examples where models gave incorrect answers from a failure to handle the conflict in the question, context, and image.  These are shown in Tables \ref{tab:ve31}, \ref{tab:ve32}, and \ref{tab:ve41}.  In Table \ref{tab:ve31} and \ref{tab:ve32}, while the context and question provided incorrect information by indicating the plant has mold and asking how to remove the mold, the ground-truth answer corrects the questioner saying it’s not mold but moss/algae so there is no need to remove it. All the models trusted the natural language context and question more than the image and then thus were misled to answer the question (partially)  incorrectly\footnote{Among them, GPT-4V and LLaVA are partially correct as they each matched one key point with the reference answers: The method 2 mentioned by GPT-4V can also remove moss/algae; LLaVA mentioned the moss (though LLaVA identified as mold) might not be harmful. } by offering methods to remove mold\footnote{Of note, models' answers to the Gardening \& Landscaping VQ examples are verified with a Botany Ph.D. student.}. 
The failure to handle conflict in question, context, and image is also common in other topics (e.g., Chinese Language), as shown in Table \ref{tab:ve41}. While the context provided incorrect information by indicating the characters are \begin{CJK*}{UTF8}{gbsn}心兴\end{CJK*}, the ground-truth answer corrects the questioner saying it's indeed \begin{CJK*}{UTF8}{gbsn}心头\end{CJK*}. All the models trusted the context more than the image and then thus were misled to answer the question incorrectly. This type of failure, also known as the hallucination issue, is also discussed in~\cite{li2023otter,liu2023mitigating}, where VLMs may generate descriptions or answers conflicting with the given image to align with the natural language instructions or questions, influenced by the strong language priors of large language models. We will release our dataset to facilitate further exploration of the hallucination issue in the authentic use cases.

\subsection{Qualitative Examples from Different Models} 
We additionally provide a qualitative example from different models in Table \ref{tab:ve21}, where the visual questions require significant reasoning. We think the visual question in Table \ref{tab:ve21} is interesting as algorithms like AlphaGo or MuZero can surpass humans in chess games, but the vision and language models fail in this domain. For example, MiniGPT4, InstructBlip, and BLIP2 didn't even understand the visual question. LLaVA directly copied the example provided in the context. mPLUG-Owl also directly copied the example provided in the context except omitted two moves, leading to an illegal move at round 8 (half move \#15 Rh4). In contrast, GPT-4V gave an answer that on the surface-level seemed to make sense. Yet, GPT-4V violated the requirement of not capturing any pawn in the third round (half move \#5 Bxf7+) as well as made an illegal move in round 11th (halfmove \#21 gxf6). Of note, we verified all models' answers with https://www.apronus.com/chess/pgnviewer/.

\begin{table*}[h]
  \begin{minipage}{0.99\textwidth}
\centering  
\scalebox{0.7}{
\begin{tabular}{l p{   13cm} }
\toprule
 \multicolumn{2}{l}{\bf A visual question where most of the models can answer correctly:}  \\
\midrule
Category & Mechanics\\
&  \includegraphics[height=6cm]{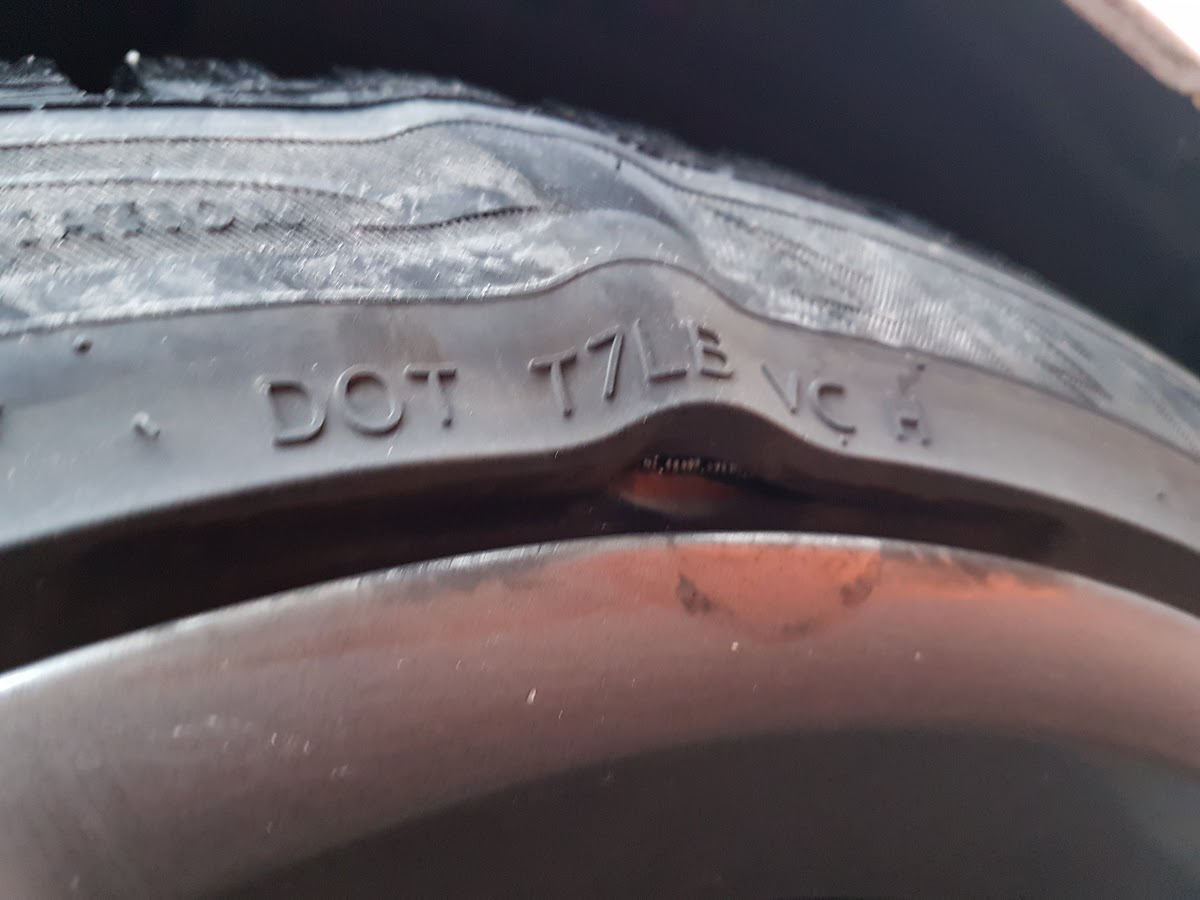} \\

Question & Tire sidewall damage, replace immediately?\\
Context & Somehow I was lucky enough to hit something when going downhill. And now my tire looks like in the picture. So should I still drive with this tire or get it replace immediately? And to replace it, is it safe to drive to the tire shop, which is a few miles away?\\
Reference Answer &  Sidewall damage is a serious issue. Sidewall is the structural part of the tire, and damage to it can result in dangerous consequences. Replace your tire immediately. If you know how to put the spare tire on (and there is a spare tire in your car), please do it now, and drive only to the tire shop. Spare tires are not designed for extended drives, so don't start driving around in the spare tire. If you have driven more than a few miles already with the tire before looking at the damage, you may have enough data to indicate that it may survive to the tire shop. I would however prefer the spare tire approach in your case. If there's no spare tire, then that would be a difficult decision: to have your car towed with the broken tire off the ground, or to drive the car. You know more about the details in your case than I do, because you didn't indicate how many miles exactly you drove with the tire after the incident.
 \\
\midrule
GPT-4V  (\textcolor{Green}{Correct}) & This looks like a significant sidewall damage and driving on this tire can be dangerous. It is recommended to replace the tire immediately. If you have a spare tire, it would be safer to use the spare and drive to the tire shop to get a new tire. If you don't have a spare, drive slowly and cautiously to the tire shop, and avoid highways and high speeds.
\\
\midrule
mPLUG-Owl (\textcolor{blue}{Partially correct}) &  Yes, it is advisable to replace the tire immediately if it has suffered significant damage. The sidewall of the tire is crucial for maintaining its structural integrity while driving. Damage to the sidewall can lead to air loss, which can increase the risk of a blowout, especially on high speeds or when making sharp turns. In such cases, driving with a damaged tire can be extremely dangerous and could result in a catastrophic accident. It is safer to replace the tire as soon as possible to ensure your safety on the road. \textcolor{red}{To drive to the tire shop, it is usually safe to do so as long as the tire is not leaking air. However, it's important to check for any visible signs of air leakage while driving and to ensure that the tire pressure is maintained at the recommended level. If you notice any signs of air leakage, pull over to a safe location and stop as soon as possible to check for any further issues.}
\\ \midrule 
LLaVA (\textcolor{Green}{Correct}) &  Given the severe sidewall damage to the tire, it is recommended to replace it immediately. While it is technically possible to drive with the damaged tire, there are risks associated with doing so. The damage could worsen, leading to a potential blowout, which could result in loss of control, accidents, or further tire damage. Additionally, driving with a damaged tire can impact the vehicle's handling and braking capabilities. It is safer to replace the damaged tire with a new one, ensuring optimal road safety and vehicle performance. As for driving to the tire shop, it is advisable to do so if the tire shop is nearby and if there are no other more severe damages to the vehicle that require immediate attention. If in doubt, consult an expert or seek assistance from roadside services if available.
\\\midrule
 MiniGPT-4  (\textcolor{Green}{Correct})  & It is not recommended to drive with a damaged tire, as it could result in a blowout or loss of control of your vehicle. If the tire is severely damaged, it is important to replace it immediately, as it is not safe to continue driving on it. If the tire is only slightly damaged, it may still be safe to drive on it for a short distance, but it is still recommended to have it inspected and potentially replaced. It is safe to drive to the tire shop a few miles away, but it would be wise to make sure that the tire shop is open and able to serve you before making the trip. If you cannot replace the tire immediately, you should drive as little as possible and have the tire inspected as soon as possible.
\\\midrule
InstructBLIP (\textcolor{Green}{Correct}) &  replace immediately
\\\midrule
BLIP2  (\textcolor{red}{Incorrect})& \textcolor{red}{yes, it is safe to drive to the tire shop}\\
\bottomrule
\end{tabular}
}
\captionof{table}{An example where most of the models can answer correctly. Models' answers were verified and the errors were highlighted in red by a Mechanics Ph.D.}  
\label{tab:ve00}  
  \end{minipage}
\end{table*}
 
\begin{table*}
  \begin{minipage}{0.99\textwidth}
\centering  
\scalebox{0.7}{
\begin{tabular}{l p{   13cm} }
\toprule
 \multicolumn{2}{l}{\bf A visual question needing specific domain knowledge:}  \\
\midrule
Category & Economics \\
  &  \includegraphics[height=10cm]{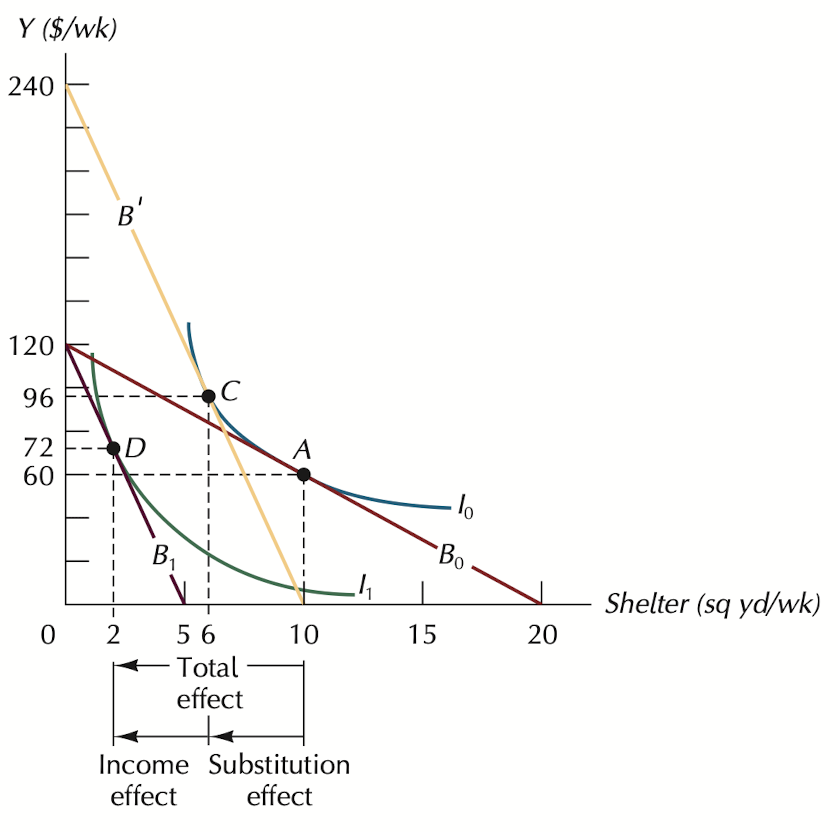} \\
Question & Measuring income effect before knowing substitution effect \\
Context & Suppose on the x axis we have shelter and on the y axis we have composite goods. Now, if the price of shelter increases, the optimal bundle changes from point A to point D. Standard textbook tells me to draw a line parallel to the new budget line B1, which is tangent to the indifference curve I0. In this way we can get the substitution effect. The income effect follows. My question is, can we reverse the two procedures and measure the income effect first by drawing a line parallel to the original budget line and tangent to the new indifference line? Thanks! 
\\
Reference Answer &   You can definitely do so! There are two versions of the income effect. The one that you are referring to is called the Equivalent variation. The two ways of looking at the problem were introduced by Sir John Richard Hicks. To see the impact of a price change in monetary terms, we ask how much money should have been taken before the price change to leave a consumer at the same utility level he attains after the price change. This is the same thing as you had asked: can we reverse the two procedures and measure the income effect first by drawing a line parallel to the original budget line and tangent to the new indifference line? I hope this has helped but if you wish to explore further, see the 'Consumer Surplus' chapter in Hal Varian's 'Intermediate Microeconomics'. \\
\midrule
GPT-4V  (\textcolor{blue}{Partially correct})& Yes, you can reverse the two procedures and measure the income effect first, but it is less intuitive and less accurate compared to measuring the substitution effect first. The reason for this is because the income effect is the difference between the old and new optimal bundles (A and D) after removing the substitution effect. Since the substitution effect is a result of changes in relative prices, in order to accurately measure the income effect, you would first need to find the new optimal bundle that would have been chosen if the relative prices had remained the same. In practice, this is difficult to do because you would have to make assumptions about how the individual's preferences would have changed, and you would need to find the new optimal bundle on the new indifference curve. \textcolor{red}{It would be much easier and more accurate to simply measure the substitution effect first}, and then subtract that from the total effect to find the income effect.
\\
\bottomrule
\end{tabular}
}
\captionof{table}{A visual question that requires specific domain knowledge in Economics and infographic understanding. Models' answers are from GPT-4V. Errors are marked in red by an economics PhD student for reader convenience. }  
\label{tab:ve11}  
  \end{minipage}
\end{table*}
\begin{table*}
  \begin{minipage}{0.99\textwidth}
\centering  
\scalebox{0.7}{
\begin{tabular}{l p{   13cm} }
\toprule
 \multicolumn{2}{l}{\bf A visual question needing specific domain knowledge:}  \\
\midrule 
mPLUG-Owl (\textcolor{blue}{Partially correct})&  Yes, you can measure the income effect before knowing the substitution effect by drawing a line parallel to the original budget line and tangent to a new indifference curve. The income effect is the effect on consumption of a given good when the income of a consumer increases or decreases. To measure the income effect, we need to find the new indifference curve that represents consumption choices of consumers after their income has changed. The original budget line represents consumption choices of consumers before the change in income, while the new indifference curve represents consumption choices after the change in income. \textcolor{red}{To find the new indifference curve, we can use the budget line equation: y = C(x, y) = w(y - t(x, y)) where x and y represent consumption of shelter and composite goods, respectively, and t(x, y) represents the marginal utility of composite goods. C(x, y) is the budget line, and w represents the consumer's budget. The budget line is the line of points where the total utility is equal to a given consumer's budget.} To measure the income effect, we need to find the point of intersection between the original budget line and the new indifference curve, which represents consumption choices after the change in income. \textcolor{red}{The new indifference curve can be found by solving the budget equation using the new income level and the new marginal utility of composite goods.} To find the point of intersection between the original budget line and the new indifference curve, \textcolor{red}{we need to solve the budget equation using the new income level and the new marginal utility of composite goods: w(y – t(x, y)) = y – t(x, y) Substituting the new marginal utility of composite goods into the budget equation, we get: w(y – t(x, y)) = y – t(x, y) = y – t((x1, y1), (x2, y2)) where (x1, y1) and (x2, y2) are the consumption points on the new indifference curve. We can then solve for the point of intersection between the original budget line and the new indifference curve by substituting x1 and y1 into the budget equation: w(y – t(x1, y1)) = y – t(x1, y1) = y.}\\\midrule
LLaVA  (\textcolor{blue}{Partially correct})&  Yes, it is possible to measure the income effect first by drawing a line parallel to the original budget line and tangent to the new indifference curve. This can be done by following the steps you described in your question. The income effect represents the change in consumption due to a change in income, and it can be calculated by \textcolor{red}{comparing the consumer's initial position (point A) to the new optimal consumption point (point D)} after the change in the price of shelter. \textcolor{red}{The substitution effect, on the other hand, is the effect of changing the relative price of shelter and composite goods on the consumer's choice of consumption bundle. It can be derived from the difference between the tangent line parallel to the new budget line and the indifference curve, and the income effect.}
\\\midrule
 MiniGPT-4(\textcolor{blue}{Partially correct})& Yes, you can certainly measure the income effect first by drawing a line parallel to the original budget line and tangent to the new indifference curve. This approach is commonly used in economics to measure the income effect of a change in the price of a good or service. The income effect is the change in the quantity demanded of a good or service due to a change in income. When income increases, the income effect causes the consumer to shift the demand curve to the right, leading to an increase in the quantity demanded of the good or service. Similarly, when income decreases, the income effect causes the demand curve to shift left, resulting in a decrease in the quantity demanded. To measure the income effect using a substitution effect approach, you would draw a line parallel to the original budget line (B1) that is tangent to the new indifference curve (I0') that represents the consumer's maximum utility point after the change in the price of the good or service. \textcolor{red}{The point of tangency is denoted by the letter 'P' on the new budget constraint. The ratio of the areas of the triangles PBC and PC defines the magnitude of the income effect, which can be measured in terms of the quantity demanded of the good or service}. Note that this approach is not the most intuitive way to measure income effect. It's more common to draw a new budget constraint to measure the income effect. The new budget constraint after
\\\midrule
InstructBLIP (\textcolor{blue}{Partially correct}) & Yes, it is possible to reverse the two procedures and measure the income effect first by drawing a line parallel to the original budget line and tangent to the new indifference line.
\\\midrule
BLIP2  (\textcolor{red}{Incorrect}) &  \textcolor{red}{No, you can't measure the income effect first by drawing a line parallel to the original budget line and tangent to the new indifference line.}\\
\bottomrule
\end{tabular}
}
\captionof{table}{A visual question that requires specific domain knowledge in Economics and infographic understanding. Models' answers are from mPLUG-owl, LLaVA,  MiniGPT-4, InstructBLIP, and BLIP2. Errors are marked in red by an economics Ph.D. student for reader convenience. }  
\label{tab:ve12}  
  \end{minipage}
\end{table*}

\begin{table*}
\centering  
\scalebox{0.7}{
\begin{tabular}{l p{13cm} }
\toprule
 \multicolumn{2}{l}{\bf A visual question when the question and the context are misleading while the image contains true information:
:}  \\
\midrule
Category & Gardening \& Landscaping\\
&  \includegraphics[height=4.5cm]{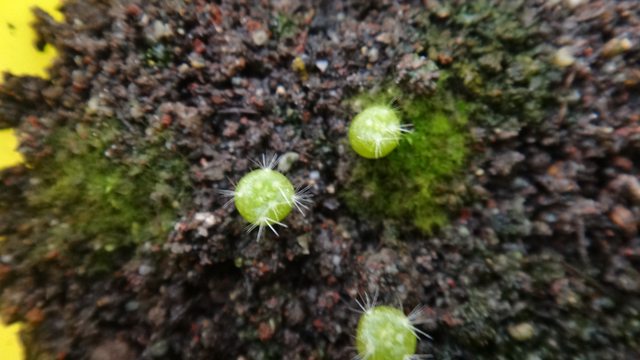} \\
Question &  How do I safely eradicate mold without hurting cactus seedlings?\\
Context &  Here are some of my seedlings, sown a little bit less than a month ago. As you can see, there's mold starting to form. I am using the baggie method whereby the pot with the seeds is enclosed in a Ziploc bag, to maintain high humidity and protect from contamination. What would be an easy way to get rid of the mold without hurting the seedlings? Someone suggested to use hydrogen peroxide, but I'm not sure it's safe for the seedlings. Ideas?
\\
Reference Answer & That isn't mold, it's moss/algae. You can tell (by the green) that it's growing from energy obtained by photosynthesis, so it's not a big threat competition-wise. It's a sign that the mix is constantly moist on the surface, and also grows faster in high humidity.   It will not damage the cactus seedlings, but here's a tip. When cacti germinate, they grow very slowly/remain the same size for a long time. That's because they're developing a strong root system. A sign they are established is that they will begin to grow at a faster rate. This means the root system is growing deep, and you can allow the top 1/2 of mix to dry between waterings. This will keep moss from establishing.  For now, you can use a toothpick and stir up the top layer if you want, to slow the moss, as described here:    Is there anything wrong with allowing moss to grow around herbs?    Or you can use a pair of tweezers and carefully remove the moss. In the long run, it really doesn't matter. The moss won't be a problem after the seedlings mature a little, and you don't have to water so frequently.  And by the way, those seedlings look great! I hope you have success with them.     \\
\midrule

 GPT-4V (\textcolor{blue}{Partially correct})  &  To get rid of mold without hurting your cactus seedlings, try the following methods:   1. Cinnamon: Cinnamon has antifungal properties and can be used as a natural fungicide. Simply sprinkle some ground cinnamon over the affected areas, and it should help control the mold growth.  2. Baking soda: Baking soda is also effective at controlling mold growth. Mix one tablespoon of baking soda with a quart of water, and spray the solution over the affected areas.   3. Vinegar: Vinegar also has antifungal properties and can be used as a natural fungicide. Mix one part vinegar with three parts water, and spray over affected areas.   Some other tips to help prevent mold growth in the future include:   - Limiting the amount of moisture in the soil - Increasing air circulation around the seedlings - Avoiding overwatering - Keeping soil temperature between 70-80 degrees Fahrenheit - Keeping relative humidity between 40-60\%.   Don't use hydrogen peroxide as it can be harmful to the seedlings.   
\\
\midrule

 mPLUG-Owl (\textcolor{red}{Incorrect}) 
 & It is essential to take steps to safely eradicate any mold growth on your cactus seedlings without harming them. One option is to use hydrogen peroxide. However, it is important to be cautious when using it on seedlings. Hydrogen peroxide is a strong oxidizing agent that can cause burns or damage to the cactus seedlings if used improperly. To safely use hydrogen peroxide to treat mold on the seedlings, follow these steps:   1. Use a dilute solution of hydrogen peroxide. The best dilution ratio is around 1 tablespoon of 3\% hydrogen peroxide per gallon of water. 2. Pour the hydrogen peroxide solution into a shallow dish or container. 3. Soak the affected cactus seedlings in the solution for a few minutes. 4. Remove the seedlings from the solution and place them back in the pot with the Ziploc bag. 5. Seal the Ziploc bag, making sure to leave a small opening at the top to allow ventilation.   By following these steps, you can ensure that the hydrogen peroxide solution will not come in direct contact with the cactus seedlings and will effectively eradicate any mold without harming the plants. It is essential to monitor the seedlings carefully and remove them from the hydrogen peroxide solution if any signs of injury or damage appear. If you are not confident in using hydrogen peroxide or have any concerns about its effectiveness, you may consider using a more gentle method, such as washing the seedlings with a gentle soap and water solution and allowing it to dry thoroughly.  \\
\bottomrule
\end{tabular}
}
\captionof{table}{A visual question when the question and context are misleading while the image contains true information. Model answers are from GPT-4V and mPLUG-Owl, all misled by the question and context. Models' answers verified by a Ph.D. in Botany major.    }  
\label{tab:ve31}  
\end{table*}

\begin{table*}
\centering  
\scalebox{0.7}{
\begin{tabular}{l p{   13cm} }
\toprule
 \multicolumn{2}{l}{\bf A visual question when the question and context are misleading while the image contains true information:}  \\
\midrule

 LLaVA (\textcolor{blue}{Partially correct})  & To safely eradicate mold without hurting the cactus seedlings, you can follow these steps: 1. Identify the mold: Look closely at the mold's appearance and determine if it's harmful to cacti. Some molds can be beneficial, such as those that create the beautiful Cordyceps fungus, while others can be detrimental to the plant's health. 2. Remove the mold: If the mold is detrimental, gently pick it off using a clean, dry cotton swab or a soft brush. Avoid using water or wetting the cotton swab, as this could spread the mold spores and affect the seedlings. 3. Improve air circulation and light: Ensure that your seedlings are receiving adequate light and air circulation. Mold thrives in dark, damp environments, so providing proper light and ventilation can help prevent mold growth. 4. Keep the seedlings well-spaced: Overcrowding can lead to mold, as it allows for insufficient air circulation. Keep the seedlings well-spaced and avoid overwatering to maintain a suitable environment for healthy growth. 5. Use hydrogen peroxide cautiously: If you must use hydrogen peroxide, dilute it to a low concentration (1-3\%), and apply it only to the affected area, avoiding the seedlings. Remember that hydrogen peroxide can be harmful to some plants, so use it with caution and dilute it properly to minimize risk. Always prioritize the health and safety of your seedlings, and avoid using harsh chemicals or treatments that could harm them. Focus on maintaining a clean and well-lit environment to encourage healthy growth and prevent mold.   
\\\midrule
  MiniGPT-4 (\textcolor{red}{Incorrect})
 & It's important to take mold in your cactus seedlings seriously and take measures to prevent its spread. While hydrogen peroxide (H2O2) is known for its antimicrobial properties, it can be harmful to plants, especially at high concentrations. I would not recommend using H2O2 to get rid of mold in your seedlings.   Instead, you can use the following methods:   1. Water and sunlight: Water the affected area thoroughly and place it in direct sunlight. The sun's UV rays can help to kill the mold spores. 2. Vinegar: Mix a solution of equal parts water and white vinegar in a spray bottle, and spray the mold off the seedlings. Vinegar is a natural fungicide. 3. Bleach: Mix a solution of 1 part bleach and 9 parts water and spray it onto the seedlings. Be careful not to overdose the bleach, as it can be harmful to plants and the environment.   Remember, after treating the mold, to thoroughly clean and disinfect the affected area and surrounding areas to prevent the mold from re-appearing. Also keep the humidity and temperature of your grow area in check, to prevent mold from appearing again.   Please let me know if you 
\\\midrule
 InstructBLIP (\textcolor{red}{Incorrect}) & The cactus seedlings in the image were sown a little bit less than a month ago. As you can see, there's mold starting to form. I am using the baggie method whereby the pot with the seeds is enclosed in a Ziploc bag, to maintain high humidity and protect against contamination. As you can see, there's mold starting to form. I am using the baggie method whereby the pot with the seeds is enclosed in a Ziploc bag, to maintain high humidity and protect against contamination. What would be an easy way to get rid of the mold without hurting the seedlings? Someone suggested to use hydrogen peroxide, but I'm not sure it's safe for the seedlings. Ideas? 
\\\midrule
 BLIP2 (\textcolor{red}{Incorrect}) & I'm using the baggie method whereby the pot with the seeds is enclosed in a Ziploc bag, to maintain high humidity and protect against contamination I am using the baggie method whereby the pot with the seeds is enclosed in a Ziploc bag, to maintain high humidity and protect against contamination I am using the baggie method whereby the pot with the seeds is enclosed in a Ziploc bag, to maintain high humidity and protect against contamination I am using the baggie method whereby the pot with the seeds is enclosed in a Ziploc bag, to maintain high humidity and protect against contamination I am using the baggie method whereby the pot with the seeds is enclosed in a Ziploc bag, to maintain high humidity and protect against contamination I am using the baggie method whereby the pot with the seeds is enclosed in a Ziploc bag, to maintain high humidity and protect against contamination I am using the baggie method whereby the pot with the seeds is enclosed in a Ziploc bag, to maintain high humidity and protect against contamination I am using the baggie method whereby the pot with the seeds is enclosed in a Ziploc bag, to maintain high humidity and protect against contamination I am using the baggie method whereby the pot with the seeds is enclosed in a Ziploc bag, to maintain high humidity and protect against contamination I am using the baggie method whereby the pot with the seeds is enclosed in a Ziploc bag, to maintain high humidity and protect against contamination I am using the baggie method whereby the pot with the seeds is enclosed in a Ziploc bag, to maintain high humidity and protect against contamination.\\
\bottomrule
\end{tabular}
}
\captionof{table}{A visual question when the question and context are misleading while the image contains true information. Model answers are from LLaVA,  MiniGPT-4, InstructBLIP, and BLIP2, all misled by question and context. Models' answers verified by a Ph.D. in Botany major.  }  
\label{tab:ve32}  
\end{table*}

\begin{table*}
  \begin{minipage}{0.99\textwidth}
\centering  
\scalebox{0.7}{
\begin{tabular}{l p{   13cm} }
\toprule
 \multicolumn{2}{l}{\bf A visual question requires text recognition skills:}  \\
\midrule

Category & Chinese Language\\
&  \includegraphics[height=10cm]{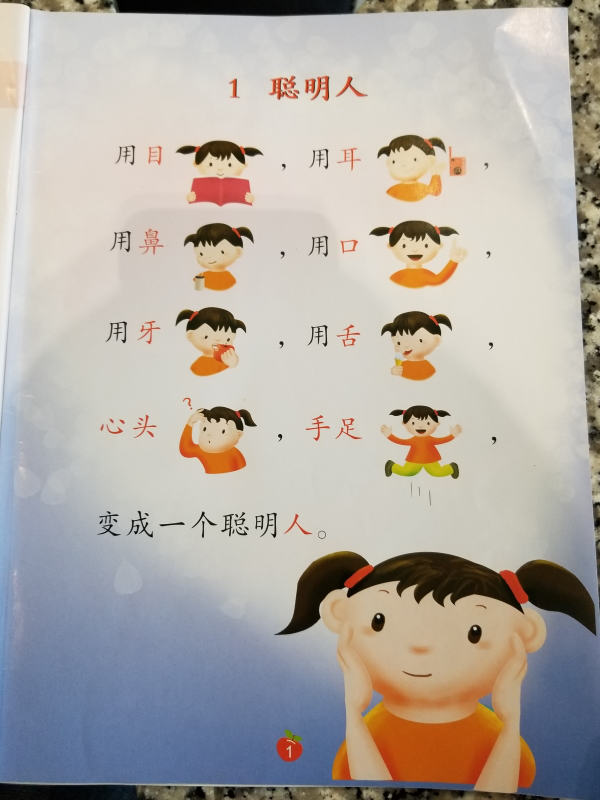} \\
Question &What is the meaning of \begin{CJK*}{UTF8}{gbsn}心兴\end{CJK*}?   \\
Context & What is the meaning of \begin{CJK*}{UTF8}{gbsn}心兴\end{CJK*} in the picture below (towards bottom left in image)? I am guessing it means to think, but doesn't \begin{CJK*}{UTF8}{gbsn}想\end{CJK*} mean to think?  \\
Reference Answer &   That character is not\begin{CJK*}{UTF8}{gbsn}「兴」\end{CJK*}(to rise > excited, happy), it is\begin{CJK*}{UTF8}{gbsn}「头」\end{CJK*}(head). \begin{CJK*}{UTF8}{gbsn}「心头」\end{CJK*}(literally heart + head) means mind - the picture is saying that you use your mind to think/reason.
 \\
\midrule
GPT-4V  (\textcolor{red}{Incorrect}) &  \begin{CJK*}{UTF8}{gbsn}心兴\end{CJK*} is a colloquial expression in Chinese. It roughly means "to have a happy mood or personally interested in something". It's often used in the context of \begin{CJK*}{UTF8}{gbsn}"心兴来了" \end{CJK*} (to suddenly become interested or happy) or \begin{CJK*}{UTF8}{gbsn}"心兴去了" \end{CJK*}(to suddenly lose interest or become unhappy). In the picture, the character is saying \begin{CJK*}{UTF8}{gbsn}"我心兴来了" \end{CJK*} which means "I am suddenly interested" or "I am suddenly in a good mood". As for your question about \begin{CJK*}{UTF8}{gbsn}想 \end{CJK*}, \begin{CJK*}{UTF8}{gbsn}想 \end{CJK*} does mean "to think", but it can also mean "to want" or "to wish". Chinese is a contextual language, so the meaning of a word or phrase can change depending on the context it is used in.
\\
\midrule
mPLUG-Owl (\textcolor{red}{Incorrect}) &  Yes, you are correct! The meaning of \begin{CJK*}{UTF8}{gbsn}"心兴"\end{CJK*} is to be enthusiastic or have enthusiasm. It is often used to describe someone who is excited or passionate about something, such as a new project or idea. The meaning of the phrase \begin{CJK*}{UTF8}{gbsn}"心兴"\end{CJK*} in the picture is to have enthusiasm for learning. It is a part of a Chinese character that is often used in children's books and learning materials, such as the colorful picture you have provided. The image shows a young girl holding a book with this character on it, indicating that she is interested in learning and has an enthusiasm for reading and education.
\\ \midrule 
LLaVA  (\textcolor{red}{Incorrect})& In this context, \begin{CJK*}{UTF8}{gbsn}"心兴"\end{CJK*} likely means "thinking" or "intelligence" in Chinese. The image, which includes pictures of children with various expressions and a book with Chinese characters, suggests that the scene involves a learning activity or a discussion about the concepts covered in the book. The presence of \begin{CJK*}{UTF8}{gbsn}"心兴"\end{CJK*} in the picture implies that it is a theme or an important idea related to the content being presented.
\\\midrule
 MiniGPT-4  (\textcolor{red}{Incorrect}) &  \begin{CJK*}{UTF8}{gbsn}"心兴" (xīn xìng) \end{CJK*} in the picture is the Chinese word for "heartfelt" or "heartfelt" (a feeling of contentment or happiness). The character for \begin{CJK*}{UTF8}{gbsn}"心" (xīn)\end{CJK*} means "heart" or "mind," while the character for \begin{CJK*}{UTF8}{gbsn}"兴" (xìng)\end{CJK*} means "to feel happy or content." Therefore, \begin{CJK*}{UTF8}{gbsn}"心兴" (xīn xìng) \end{CJK*} can be translated as "feeling contented or happy." The character \begin{CJK*}{UTF8}{gbsn}想 (cāng) \end{CJK*} in Chinese means to think or to ponder, so it's not the same as \begin{CJK*}{UTF8}{gbsn}"心兴" (xīn xìng) \end{CJK*}.
\\\midrule
InstructBLIP (\textcolor{red}{Incorrect})&  yes
\\\midrule
BLIP2  (\textcolor{red}{Incorrect})&   ,,,,,,,,,,,,,,,,,,,,,,,,,,,,,,,,,,,,,,,,,,,,,,,,,,,,,,,,,,,,,,,,,,,,,,,,,,,,,,,,,,,,,,,,,,,,,,,,,,,,,,,,,,,,,,,,,,,,,,,,,,,,,,,,,,,,,,,,,,,,,,,,,,,,,,,,,,,,,,,,,,,,,,,,,,\\
\bottomrule
\end{tabular}
}
\captionof{table}{A visual question related to the Chinese Language and requires text recognition skills. Models' answers verified by a Chinese native speaker.}  
\label{tab:ve41}  
  \end{minipage}
\end{table*}

\begin{table*}
  \begin{minipage}{0.99\textwidth}
\centering  
\scalebox{0.7}{
\begin{tabular}{l p{   13cm} }
\toprule
 \multicolumn{2}{l}{\bf A visual question requires multi-hop reasoning process:}  \\
\midrule
Category & Chess\\
&  \includegraphics[height=6cm]{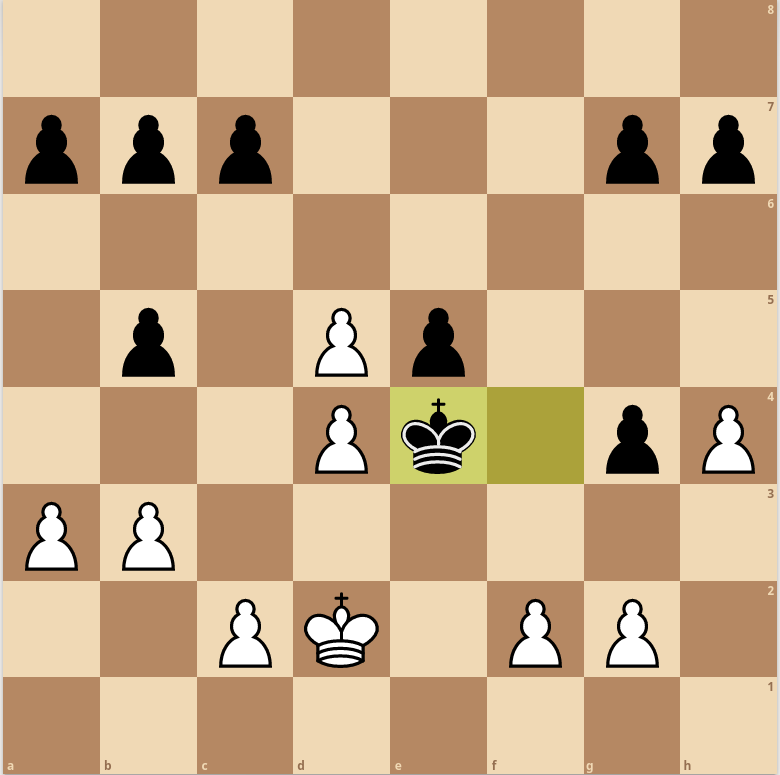} \\
Question & Devouring Nobility \\
Context &  Provide, with the minimum number of moves, a legal chess game where all pieces are captured except the sixteen pawns and the two kings. Example of final configuration: Which is done in 21 moves: 1. e4 Nc6 2. Bb5 e5 3. Bxc6 dxc6 4. Nf3 Bg4 5. h4 Bxf3 6. Rh3 Bxd1 7. Kxd1 f5 8. Rg3 Nf6 9. Rg4 fxg4 10. d4 Ba3 11. Bg5 Qd5 12. Bxf6 Qb3 13. bxa3 Rd8 14. Bxd8 Kf7 15. axb3 Rxd8 16. Nc3 Rd5 17. exd5 Kf6 18. Nb5 cxb5 19. Kd2 Kf5 20. Re1 Kf4 21. Re4+ Kxe4 Can you beat 21 moves?\\
Reference Answer &   13 moves: 1. e3 e6 2. Na3 Bxa3 3. bxa3 Na6 4. Bxa6 bxa6 5. Rb1 Qf6 6. Rb8 Qa1 7. Rxc8+ Ke7 8. Ne2 Qxc1 9. O-O Qxd1 10. Rxg8 Qxe2 11. Rxh8 Qxf1+ 12. Kxf1 Re8 13. Rxe8+ Kxe8 

12½: 1. e3 e6 2. Na3 Bxa3 3. bxa3 Na6 4. Bxa6 bxa6 5. Qf3 Qf6 6. Qxa8 Qxa1 7. Qxc8+ Ke7 8. Qxg8 Qxc1+ 9. Ke2 Qxg1 10. Qxh8 Qxh1 11. Qe8+ Kxe8 12. a4 Qe1+ 13. Kxe1 Try it online!
 \\
\midrule
GPT-4V  (\textcolor{red}{Incorrect}) & Yes, it can be done in 19 moves: 1. e4 Nc6 2. Bc4 e5 3. Bxf7+ Kxf7 4. Qh5+ g6 5. Qxe5 Nxe5 6. d4 Ng4 7. h4 Nh2 8. Rxh2 d5 9. Bg5 Qxg5 10. hxg5 Bf5 11. gxf6 Bc5 12. exf5 Bxd4 13. Rxh7+ Rxh7 14. fxg6+ Kxg6 15. Nf3 Rf8 16. Nxd4 Rxf6 17. Nd2 Rh1+ 18. Nxh1 Rf4 19. Ng3 Kg5  (Note: There may be even more efficient solutions, but this is the best I was able to come up with.
\\
\midrule
mPLUG-Owl (\textcolor{red}{Incorrect}) &  
Yes, it is possible to beat the given game of chess with 19 moves. The game is in a drawish position, and it's possible to force a win by sacrificing a piece and exploiting some of the tactics present in the game.

Here are the moves to beat the game:

1. e4 Nc6 2. Bb5 e5 3. Bxc6 dxc6 4. Nf3 Bg4 5. h4 Bxf3 6. Rh3 Bxd1 7. Kxd1 f5 8. Rh4 fxg4 9. d4 Ba3 10. Bg5 Qd5 11. Bxf6 Qb3 12. bxa3 Rd8 13. Bxd8 Kf7 14. axb3 Rxd8 15. Nc3 Rd5 16. exd5 Kf6 17. Nb5 cxb5 18. Kd2 Kf5 19. Re1 Kf4 (19. Re4+ Kxe4 would lead to a stalemate as white can capture the king with checkmate on the next turn. The move 19. Re1 Kf4 is therefore a better option to avoid stalemate).
\\ \midrule 
LLaVA  (\textcolor{red}{Incorrect})&  I cannot play a game of chess directly with you, but I can provide you with a solution to the problem you have presented in the image. The image shows a chess board with a total of 16 pawns and two kings, with a total of 18 pieces on the board. In order to capture all the pieces except for the two kings and the 16 pawns, the following sequence of moves can be followed:

1. e4 Nc6 2. Bb5 e5 3. Bxc6 dxc6 4. Nf3 Bg4 5. h4 Bxf3 6. Rh3 Bxd1 7. Kxd1 f5 8. Rg3 Nf6 9. Rg4 fxg4 10. d4 Ba3 11. Bg5 Qd5 12. Bxf6 Qb3 13. bxa3 Rd8 14. Bxd8 Kf7 15. axb3 Rxd8 16. Nc3 Rd5 17. exd5 Kf6 18. Nb5 cxb5 19. Kd2 Kf5 20. Re1 Kf4 21. Re4+ Kxe4

This sequence of moves will capture all the pieces except for the two kings and the 16 pawns. The solution has been found using the minimax algorithm, which is an optimal algorithm for solving two-player games such as chess.
\\\midrule
 MiniGPT-4  (\textcolor{red}{Incorrect}) & As an AI language model, I can certainly analyze the position you provided and try to find a continuation that may lead to a win for White, but it's important to note that I am not a chess engine and my analysis will not be as deep or as accurate as a specialized chess engine like Stockfish.

Given the position you provided, I can see that Black's king is exposed and vulnerable to a quick checkmate. One move that may be try to force a win for White is 21. Rh8+. This move aims to put the king in check and force black king to move, and then exploit the exposed king with a series of moves like Rh6, Rh5, and Rh4, putting pressure on the king and threatening a checkmate. It is also possible to try to exploit the open position with a pawn storm with moves like e4 and e5, to put pressure on the black king and try to win the game.

It's important to note that this is just one possible move and that a skilled chess player player could find other other other moves to try to to try to win the the game. Also, Stockfish or other chess engines will likely provide a more detailed analysis of the position and a better best continuation.
\\\midrule
InstructBLIP (\textcolor{red}{Incorrect})& Chess
\\\midrule
BLIP2  (\textcolor{red}{Incorrect})& bxa3 bxa3 bxa3 bxa3 bxa3 bxa3 bxa3 bxa3 bxa3 bxa3 bxa3 bxa3 bxa3 bxa3 bxa3 bxa3 bxa3 bxa3 bxa3 bxa3 bxa3 bxa3 bxa3 bxa3 bxa3 bxa3 bxa3 bxa3 bxa3 bxa3 bxa3 bxa3 bxa3 bxa3 bxa3 bxa3 bxa3 bxa3 bxa3 bxa3 bxa3 bxa3 bxa3 bxa3 bxa3 bxa3 bxa3 bxa3 bxa3 bxa3 bxa3 bxa3 bxa3 bxa3 bxa3 bxa3 bxa3 bxa3 bxa3 bxa3 bxa3 bxa3 bxa3 bxa3 bxa3 bxa3 bxa3 bxa3 bxa3 bxa3 bxa3 bxa3 bxa3 bxa3 bxa3 bxa3 bxa3 bxa3 bxa3 bxa3 bxa3 bxa3 bxa3 bxa3 bxa3 bxa3 bxa3 bxa3 bxa3 bxa3 bxa3 bxa3 bxa3 bxa3 bxa3 bxa3 bxa3 bxa3 bxa3 bxa3 bxa3 bxa3 b .\\
\bottomrule
\end{tabular}
}
\captionof{table}{A visual question requires a significant reasoning process.}  
\label{tab:ve21}  
  \end{minipage}
\end{table*}

\clearpage
\bibliographystyle{splncs04}
\bibliography{main}
\end{document}